\documentclass[11pt]{article}

\usepackage[final]{acl}

\usepackage{times}
\usepackage{latexsym}

\usepackage[T1]{fontenc}

\usepackage[utf8]{inputenc}

\usepackage{microtype}

\usepackage{inconsolata}
\usepackage{amsmath}
\usepackage{amssymb}
\usepackage{booktabs}
\usepackage{adjustbox}
\usepackage{multirow}
\usepackage{siunitx}
\usepackage{graphicx}
\usepackage{hyperref}
\usepackage{subcaption}
\usepackage[capitalise,noabbrev]{cleveref}
\title{Task-Aware LLM Routing with Multi-Level Task-Profile-Guided Data Synthesis for Cold-Start Scenarios}

\author{
Hui Liu\textsuperscript{1},
Bin Zou\textsuperscript{2},
Kecheng Chen\textsuperscript{1},
Jie Liu\textsuperscript{1},
Wenya Wang\textsuperscript{3},
Haoliang Li\textsuperscript{1}\\[0.6em]
\textsuperscript{1}City University of Hong Kong \quad
\textsuperscript{2} University of Hong Kong  \quad
\textsuperscript{3}Nanyang Technological University
}

\begin{document}
\maketitle
\begin{abstract}
Large language models (LLMs) exhibit substantial variability in performance and computational cost across tasks and queries, motivating routing systems that select models to meet user-specific cost–performance trade-offs. However, existing routers generalize poorly in cold-start scenarios where in-domain training data is unavailable. We address this limitation with a multi-level task-profile–guided data synthesis framework that constructs a hierarchical task taxonomy and produces diverse question–answer pairs to approximate the test-time query distribution. Building on this, we introduce TRouter, a task-type–aware router approach that models query-conditioned cost and performance via latent task-type variables, with prior regularization derived from the synthesized task taxonomy. This design enhances TRouter's routing utility under both cold-start and in-domain settings. Across multiple benchmarks, we show that our synthesis framework alleviates cold-start issues and that TRouter delivers effective LLM routing. The code is publicly available at this link\footnote{\url{https://github.com/less-and-less-bugs/ColdStartLLMRouter}}.
\end{abstract}

\section{Introduction}
Large language models (LLMs) have been widely adopted by consumers and enterprises~\citep {appelmccrorytamkin2025geoapi}, yet their performance and cost vary substantially across tasks and queries due to parameter size and reasoning length differences~\citep{hu2024routerbench}. Moreover, users also have different preferences for performance and cost~\citep{frugalgpt}. For example, a startup deploying a customer chatbot may prioritize cost-effectiveness, whereas a research lab analyzing complex literature may favor high-performance models despite higher costs. This diversity motivates the LLM routing problem of assigning optimal models from a candidate pool to each query while aligning with user preferences~\citep{graphrouter, lu2023routing}.
\begin{figure}[t]
\centering
\includegraphics[width=\linewidth]{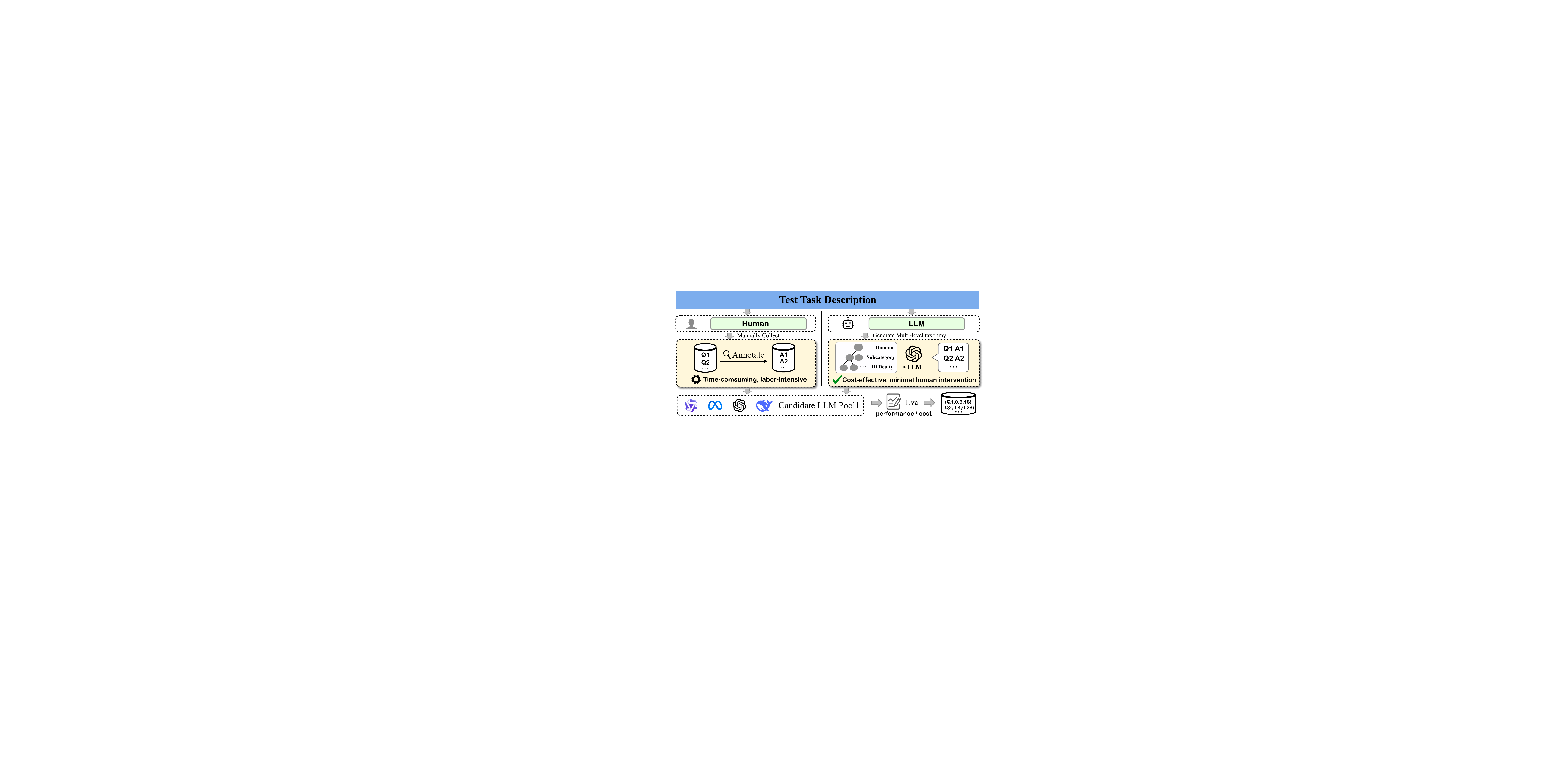}
\caption{Comparison between the traditional data preparation pipeline and our proposed LLM-based data synthesis approach for the LLM router training.}
\vspace{-10pt}
\label{fig:fig1}
\end{figure}

As shown in Fig~\ref{fig:fig1}, most existing approaches assume access to training datasets (in-domain setting) with query-answer (QA) pairs and evaluate all candidate LLMs to obtain performance and cost metrics. They then train lightweight routers, either classifiers or regressors~\citep{devlin2019bert} to select models. Classification-based approaches~\citep{ong2024routellm,zhuang2024embedllm,metallm,routerdc} label optimal LLMs according to user preferences and output the selected model during testing, while regression-based methods~\citep{capabilityrouter, carrot, mohammadshahi2024routoo,vsakota2024fly} predict performance and cost and select models via utility maximization, enabling flexible specifications such as batch routing~\citep{omnirouter}.

However, real-world deployments frequently face cold-start scenarios without resources to collect labeled training data~\citep{wang2025knowledge, liu2025enhancing}, especially for individual users and early-stage products. Moreover, pre-trained routers often generalize poorly at deployment, likely due to domain shifts between training and test-time user inputs~\citep{routerr1}. As shown in Table~\ref{tab:maintraditionmetric}, both classification and regression-based methods (e.g., RouterDC and MetricRouter) exhibit limited robustness under cross-domain settings compared to a simple rule-based alternative (Adaptive LLM) that selects more powerful models as user cost tolerance increases. Additionally, while routing queries directly with separate LLMs may seem feasible, accurately characterizing each candidate model’s capabilities remains difficult~\citep{chang2024survey, yehudai2025survey}, making this approach unsuitable for cold-start scenarios.

To address the cold-start problem in LLM routing, we propose a multi-level task-profile-guided data synthesis framework that generates diverse question-answer pairs approximating the test-time query distribution. Drawing upon recent advances in LLM evaluation and deployment practices~\citep{chang2024survey,shao2024deepseekmath}, we observe that both computational cost and model performance are intrinsically linked to task category and difficulty, further validated in Appendix~\ref{anaonDsyn}. Our framework employs a Task Type Generator that, when seeded with domain-relevant information, iteratively constructs a hierarchical task taxonomy spanning domain, subcategory, and difficulty dimensions. This hierarchical mechanism enables fine-grained control and enhanced sampling efficiency, while a Task Type Quality Evaluator ensures taxonomic cohesion and diversity throughout the generation process. By conceptualizing each difficulty-specific category as a distinct task profile, our Question-Answer Pair Generator produces non-redundant QA pairs for each profile, which candidate LLMs subsequently evaluate to generate routing training data for cold-start scenarios.  

Departing from conventional approaches that directly model cost and performance from raw query features, we introduce TRouter, a task-type-aware LLM routing system that augments regression-based routing through the incorporation of latent task-type variables and employs a prior distribution over the synthesized task taxonomy as a regularization. TRouter effectively captures latent task semantic structures that transcend surface-level query characteristics, achieving superior routing performance and enhanced robustness across cold-start and in-domain configurations. Extensive experiments across multiple LLM pools and evaluation protocols show that our synthesis framework mitigates cold-start limitations, while TRouter exploits task-specific information to yield substantial improvements in routing utility. The contribution of this work is as follows: 
\begin{itemize}
    \item To the best of our knowledge, we are the first to identify the cold-start problem in LLM routing and introduce a multi-level task-profile–guided data synthesis framework that approximates the test-time query distribution.
    \item We propose TRouter, a task-type–aware router that integrates latent task-type variables and a prior over the synthesized task taxonomy into the regression-based paradigm, improving routing utility. 
    \item We validate the effectiveness of the synthesis framework and TRouter across multiple LLM pools and evaluation protocols in both cold-start and in-domain settings
\end{itemize}


\section{RELATED WORKS}
\paragraph{LLM Routing.} LLM routing aims to optimize the allocation of candidate LLMs for input queries under user-defined preferences. While early work focuses on performance maximization~\citep{jiang2023llm,shnitzer2023large}, recent approaches emphasize performance-cost trade-offs as model capabilities improve and call costs rise~\citep{zhang2023ecoassistant,hybridllm}. Using two main paradigms, these methods evaluate candidate LLMs on training queries and train a small model, like BERT, as a router. Classification-based approaches~\citep{routerdc,srivatsa2024harnessing} annotate optimal LLMs for training queries and learn to predict the best model. For example, GraphRouter~\citep{graphrouter} formulates routing as edge classification over heterogeneous graphs modeling query-model relationships. Regression-based methods~\citep{carrot,omnirouter} predict performance and cost metrics for query-model pairs and then optimize user-defined utility functions at inference. Our work adopts the regression paradigm because it can enable batch routing with model quotas~\citep{cheng2024enabling} and support fine-grained preference control, whereas classification methods typically operate per-query. Additionally, alternative approaches directly fine-tune LLMs for model selection, potentially achieving higher accuracy but increasing inference costs and latency~\citep{routerr1,capabilityrouter,automix}. At the same time, cascade-based methods assign fixed LLM sequences to the whole test set, limiting their performance in query-level routing~\citep{frugalgpt, C2MAB-V}.
\paragraph{Data Synthesis using LLMs.}  LLM-based data synthesis has substantially reduced annotation costs, enabling scalable construction of large-scale training datasets. Recent work~\citep{tan2024large} identifies three core principles for effective synthesis: diversity, achieved via conditional prompting and multi-step generation to broaden topical and stylistic coverage; quality, maintained through filtering, label refinement, and cross-consistency checks to limit hallucinations and noise; and domain specificity, incorporating external knowledge and task-tailored prompts to inject expertise. However, empirical studies~\citep{yang2025measuring, shumailov2024ai} demonstrate that synthesis pipelines are highly customized, with different applications requiring distinct approaches. To our knowledge, no existing work has explored data synthesis methods to approximate test-time query distributions for LLM routing training.


\section{METHODOLOGY}
\begin{figure*}[t]
\centering
\includegraphics[width=\linewidth]{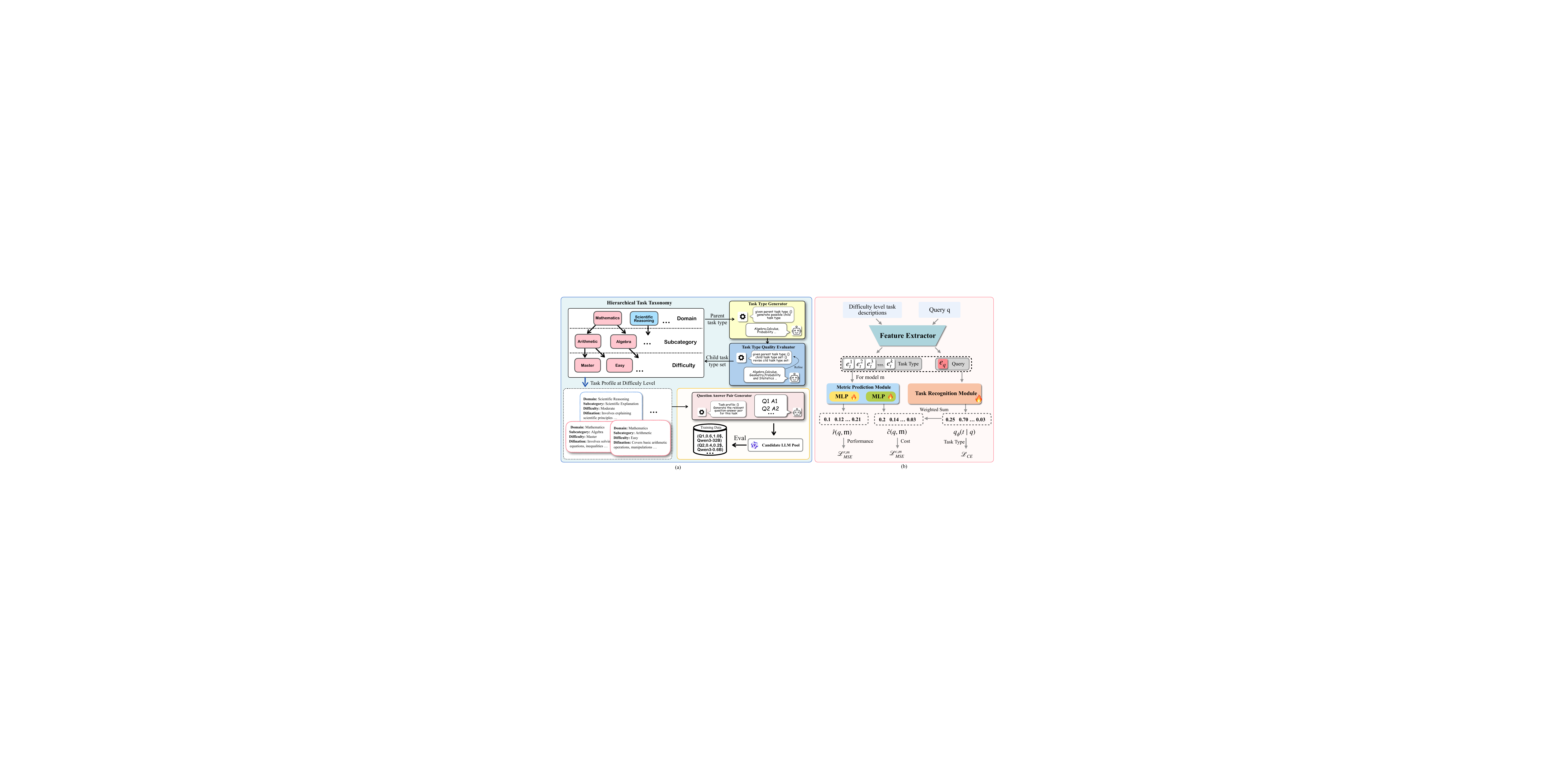}
\caption{(a) Overview of our proposed task-profile-guided data synthesis framework. The task type generator and quality evaluator collaboratively construct a hierarchical task taxonomy, while the question-answer pair generator produces diverse QA pairs based on task profiles from difficulty-level task types. (b) Overview of our proposed task-type–aware LLM routing method, introducing a latent task-type variable into cost and performance prediction.}
\vspace{-5pt}
\label{fig:fig2}
\end{figure*}

To mitigate cold-start challenges in practical LLM routing, we propose a multi-level task-profile–guided data synthesis framework that generates diverse question–answer pairs to approximate realistic query distributions in Sec.~\ref {sec.3.2}. Leveraging the synthesized task taxonomy, we further introduce TRouter, a task-type–aware router that treats task type as a latent variable to predict performance and cost in Sec.~\ref {sec.3.3}. Fig. \ref{fig:fig2} provides an overview of our proposed data synthesis strategy and TRouter.
\subsection{Problem formulation}
Let $\mathcal{M} = \{m_1, \ldots, m_M\}$ denote a set of $M$ candidate LLMs. Following \citet{graphrouter}, for a user query $q$, LLM routing aims to select a model $ m^* \in \mathcal{M}$ to maximize a utility function as below:
\begin{equation}
\small
\label{qe:utility}
\begin{aligned}
m^* &= \arg\max_{m \in \mathcal{M}}  U(m, q) = \mu_r \cdot \mathbf{r}(m, q) - \mu_c \cdot \mathbf{c}(m, q) \\
& \text{s.t.} \quad \mu_r + \mu_c = 1, 
\quad \mu_r, \mu_c \geq 0
\end{aligned}
\end{equation}
where $\mathbf{r}(\cdot)$ and $\mathbf{c}(\cdot)$ are learnable performance and cost functions, and $\mu_r, \mu_c$ represent user preferences over the performance–cost trade-off.
Performance evaluation may vary depending on task types (e.g., accuracy and F1), while cost is determined by input/output token numbers and model-specific per-token pricing. In a given test scenario with test set $\mathcal{D}_{\text{test}}$, prior work often assumes access to an in-domain labeled training set and evaluates all models in $\mathcal{M}$ to collect tuples $\{q_i, c_{i,j}, r_{i,j}, m_j\}$, where $c_{i,j}$ and $r_{i,j}$ are the observed cost and performance of model $m_j$ on query $q_i$. These tuples constitute the final router training data $\mathcal{D}_{\text{train}}$.
\subsection{Task-profile-guided Data Synthesis} 
\label{sec.3.2}
As obtaining sufficient manually annotated data is challenging in the cold-start setting, we propose a multi-level task-profile-guided data synthesis framework, comprising task type generator, task type quality evaluator, and question-answer pair generator. 
Each component is detailed as follows.
\paragraph{Task Type Generator.} We adopt a three-level task taxonomy from general to fine-grained, comprising domain, subcategory, and difficulty. A short name, a definition, and an illustrative example are generated to specify each task type. Beginning with manually authored domain descriptions, this generator recursively expands the taxonomy by producing multiple distinct, non-overlapping child types per step. Parent-type definitions are used as conditional prompts to promote diversity and relevance. To avoid excessive granularity, we enforce maximum node counts via prompt constraints.
\paragraph{Task Type Quality Evaluator.} After the task type generator proposes a set of candidate child task types for a given parent type, the task type quality evaluator conducts a self-critique~\citep{shao2024deepseekmath} against predefined criteria, such as minimal redundancy, specificity, and completeness. The evaluator first shuffles the current set and ingests it as input, then decides whether revisions are warranted. If so, it iteratively refines the set. This process repeats until the evaluator determines that no further modifications are necessary across multiple consecutive assessment rounds. At that point, the candidate set is finalized as the child task type set for the corresponding parent type.
\paragraph{Question-Answer Pair Generator.} After constructing the hierarchical task taxonomy, we use the short name and description of each difficulty-level task type, together with the descriptions of its parent task types, to form a task profile. For each profile, our objective is to generate $N_d$ QA pairs. The QA pair generator produces QA pairs in batches and applies a sentence-transformer model to compute the semantic similarity between newly generated pairs and the existing collection. Any QA pair whose maximum similarity to an existing pair exceeds 0.9 is discarded to filter out near-duplicate ones. This batched generation and filtering process is repeated iteratively until $N_d$ diverse QA pairs are obtained for each task profile.

Overall, our proposed data synthesis framework requires only domain descriptions as input to guide LLMs in constructing a synthetic QA dataset, thus obviating manual data collection and annotation. Similar to $\mathcal{D}_{\text{train}}$, all candidate LLMs in $\mathcal{M}$ are evaluated on the QA dataset to collect routing training data, denoted as $\mathcal{D}_\text{syn}$ in a cold-start setting.
\subsection{Task-type–aware Routing Framework}
\label{sec.3.3}
\subsubsection{Theoretical Analysis}
\label{sec.3.3.1}
Let $h \in \mathcal{H} = \{r, c\}$ denote an evaluation metric, where $r$ and $c$ represent performance and cost, respectively. In contrast to prior routing methods that directly model the conditional distributions of cost and performance for query-model pairs, i.e., $p(h \mid q,m)$, TRouter introduces an unobserved latent task variable $t \in \mathcal{T}$, where $\mathcal{T}$ represents the task-type space. The conditional distribution is then decomposed as:
\begin{equation}
\small
p(h \mid q, m) = \sum_{h \in \mathcal{T}} p(h \mid q, m, t) p(t \mid q,m).
\end{equation}
By leveraging $t$ as the intermediate representation across different metrics, TRouter disentangles the impact of task semantics from other input features.

To simplify the probability distribution, we adopt two conditional independence assumptions: (i) performance is independent of the query given the task type and model, and (ii) task type is independent of the model given the query, which are denoted as:
\begin{equation}
\small
p(h \mid t, q, m) = p(h \mid t, m), \quad p(t \mid q, m) = p(t \mid q).
\label{eq:3}
\end{equation}
Under these assumptions, the conditional distribution $p(h \mid q, m)$ can be reformulated as:
\begin{equation}
\small
\label{eq:4}
p(h \mid q, m) = \sum_{t \in \mathcal{T}} p(h \mid t, m)p(t \mid q).
\end{equation}
Intuitively, the query $q$ induces a posterior $p(t \mid q)$ over task types, and each task type $t$ then specifies a task-model-pair dependent metric conditional distribution $p(h \mid t,m)$. 

To train the metric predictor $p_\theta(h \mid q, m)$, given a training corpus $\mathcal{D}$ (e.g., $\mathcal{D}_{\text{syn}}$ and $\mathcal{D}_{\text{train}}$), we maximize the log-likelihood through:

\begin{small}
\begin{equation}
\begin{aligned}
\mathcal{L}
&= \sum_{(q,h,m) \in \mathcal{D}} \log p_\theta(h \mid q, m) \\
&= \sum_{(q,h,m) \in \mathcal{D}} \log \sum_{t \in \mathcal{T}} p_{\theta}(h \mid t, m)p(t \mid q),
\end{aligned}
\label{eq:training-objective}
\end{equation}
\end{small}where $p(t \mid q)$ is the prior over task types and $\theta$ denotes learnable parameters. However, directly maximizing such a likelihood is difficult due to marginalization over the latent $t$ and because $t$ is unobserved at deployment. We therefore introduce a variational posterior $q_{\phi}(t\mid q)$ to approximate $p(t \mid q)$ and optimize the evidence lower bound (ELBO)~\citep{kingma2013auto} as follows

\begin{small}
\begin{multline}
\log p_\theta(h \mid q, m) \geq \mathbb{E}_{q_{\phi}(t \mid q)} \left[\log p_{\theta}(h \mid t, m)\right]\\
- \text{KL}\left(q_{\phi}(t \mid q) \,\|\, p(t \mid q)\right),
\label{eq:elbo-r}
\end{multline}
\end{small}where $\phi$ denotes learnable parameters. The first term is a reconstruction term that encourages task representations predictive of metric $h$ for model $m$; the second term regularizes the variational posterior toward the prior, improving calibration and mitigating overfitting. A complete derivation of Eq~\eqref{eq:elbo-r} is detailed in Appendix~\ref{app:ssupptheoana}.
\subsubsection{Model Implementation}
We instantiate our probabilistic framework with a task recognition module ($q_{\phi}(t \mid q)$) and multiple metric prediction modules ($p_{\theta}(h \mid t, m)$). 
We define the full set of task types at the difficulty level of the task taxonomy introduced in Sec. \ref{sec.3.2} as $\mathcal{T}$ with $|\mathcal{T}|=K$. For a given query $q$, while it may belong to multiple task types in $\mathcal{T}$, we assign a single ground-truth label by selecting the task type corresponding to the query's annotated difficulty level. This label is represented as a one-hot vector $y \in \mathbb{R}^K$ and serves as prior ($p(t \mid q)$) for the recognition module. Both the query $q$ and the textual definition of task types in $\mathcal{T}$ are encoded using a pretrained sentence encoder. The resulting representations are then passed through two-layer multilayer perceptrons (MLPs) to produce a query embedding $e_q$ and task-type embeddings $\{e_t^i\}$  where $i\in[1,K]$.
\paragraph{Task recognition module.} For the query $q$, we concatenate its embedding and all task embeddings $\{e_t^i\}$  and pass them into an MLP, followed by a softmax function with a temperature parameter $\tau$  to obtain the predicted task type distribution $l \in \mathbb{R}^K$ as $q_{\phi}(t \mid q)$. Then we train this module using cross-entropy loss over the training set $\mathcal{D}$ as follows:
\begin{equation}
\small
\mathcal{L}_{\text{CE}} = -\sum_{q\in \mathcal{D}}y \log q_{\phi}(t \mid q),
\end{equation}
enforcing the optimization of KL-term in Eq. \eqref{eq:elbo-r}.
\paragraph{Metric prediction module.} For each pair of metric $h$ and model $m \in \mathcal{M}$, we predict the metric by computing the weighted combination of task-specific predictions using the predicted task distribution $l$ from the recognition module as follows:
\begin{equation}
\small
\hat{h}(q, M) = \sum_{i \in [1,K]} l_i\cdot \sigma\left(\text{MLP}_{h,m}(e_t^i)\right),
\end{equation}
where $\sigma$ denotes the sigmoid function, ensuring normalized outputs, and $l_i$ represents the possibility that $q$ belongs to $i$-th task type. We train this module with MSE loss between predicted and observed metric $h(q, M)$ as follows:
\begin{equation}
\small
\mathcal{L}_{\text{MSE}}^{h,m} = \sum_{q\in \mathcal{D}}\left(\hat{h}(q, m) - h(q, m)\right)^2,
\end{equation}
which serves as reconstruction term in Eq. \eqref{eq:elbo-r}.
Then, we combine both losses to form the final training objective over the training set $\mathcal{D}$ as follows:
\begin{equation}
\small
\mathcal{L} = \mathcal{L}_{\text{CE}} + \frac{1}{|\mathcal{M}||\mathcal{H}|} \sum_{m \in \mathcal{M}} \sum_{h \in \mathcal{H}}\mathcal{L}_{\text{MSE}}^{h,m}.
\end{equation}
During inference, for each query, we compute predicted metrics for each model via marginalization over the task space $\mathcal{T}$ and select the model that maximizes the utility function $U(m, q)$ in Eq. \eqref{qe:utility}.

\section{Experiments}
\subsection{Experimental Setting}
\begin{table}[!t]
\centering
\caption{Overview of Datasets.}
\label{tab:dataset}
\begin{adjustbox}{width=0.85\linewidth}
\begin{tabular}{l l l r}
\toprule
\textbf{Dataset} & \textbf{Task Type} & \textbf{Metric} & \textbf{Cases} \\
\midrule
Alpaca~\citep{taori2023alpaca}      & Hybrid QA               & F1        & 2000 \\
GSM8K~\citep{cobbe2021training}       & Reasoning               & Accuracy  & 2000 \\
SQuAD~\citep{rajpurkar2016squad}       & Reading Comprehension   & F1        & 2000 \\
Multi-News~\citep{fabbri2019multi}  & Summary                 & F1        & 2000 \\
\bottomrule
\end{tabular}
\end{adjustbox}
\vspace{-10pt}
\end{table}
\begin{table}[!t]
\centering
\caption{Overview of open-source and commercial LLMs, their sizes, and costs per million tokens.}
\label{tab:modelstat}
\begin{adjustbox}{width=0.85\linewidth}
\begin{tabular}{llrr}
\toprule
\textbf{Model} & \textbf{Size} & \textbf{Input (¥/M)} & \textbf{Output (¥/M)} \\
\midrule
\multicolumn{4}{l}{\textbf{Open-source}} \\
\midrule
Qwen3-235B-A22B     & 235B & 2.00 & 8.00 \\
Qwen3-32B           & 32B  & 2.00 & 8.00 \\
Qwen3-14B           & 14B  & 1.00 & 4.00 \\
Qwen3-8B            & 8B   & 0.50 & 2.00 \\
Qwen3-1.7B          & 1.7B & 0.30 & 1.20 \\
Qwen3-0.6B          & 0.6B & 0.30 & 1.20 \\
\midrule
\multicolumn{4}{l}{\textbf{Commercial}} \\
\midrule
Gemini-2.5-Flash              & --   & 2.14 & 17.86 \\
Gemini-2.5-Flash-Lite         & --   & 0.71 & 2.86  \\
Gemini-2.0-Flash       & -- & 0.71 & 2.86  \\
Gemini-2.0-Flash-Lite-Preview & --   & 0.54 & 2.14  \\
Doubao-Seed-1.6-Flash         & --   & 0.15 & 1.50  \\
\bottomrule
\end{tabular}
\end{adjustbox}
\vspace{-10pt}
\end{table}
\paragraph{Datasets and Candidate LLMs.} Following recent work~\citep{graphrouter}, we select data from four distinct tasks, with their statistics in Table~\ref{tab:dataset}. We further validate the versatility of our approach on four additional tasks in Appendix~\ref{app:suppexe}. Instead of using outdated LLMs, to better reflect realistic deployment demands, we employ more advanced  Qwen 3 models of varying sizes in our primary experiments and conduct supplementary experiments using a group of proprietary models in Table~\ref{tab:modelstat}.
\paragraph{Data Prepossessing.}
We aggregate all sampled data from the four tasks. For each query, we utilize the candidate LLMs in Table~\ref{tab:modelstat} to generate corresponding responses. While prior work often assumes that the evaluation protocol for the test scenario is available, we argue that such assumptions do not always hold in real-world settings. Therefore, to evaluate response quality, in addition to employing traditional metrics in Table~\ref{tab:modelstat}, we apply an LLM-as-a-judge framework~\citep{gu2024survey}, where a designated LLM evaluates responses based on predefined rules, serving as a proxy for ground-truth performance. Moreover, the cost of each model is computed by summing the number of input and output tokens, weighted by the corresponding price.  Upon constructing the multi-task interaction dataset, we divide it into training, validation, and test subsets using a 7:1:2 ratio. 

For the synthetic data generated via the task-profile guided data synthesis method (Sec.~\ref{sec.3.2}), we estimate performance using the LLM-as-a-judge and calculate cost following the same procedure.

\paragraph{Metrics.} Since users may prioritize performance and cost differently, we define three user preference scenarios, including Cost First, Balanced, and Performance First. Specifically, we assign weights $(\mu_r, \mu_c)$ of $(0.2, 0.8)$, $(0.5, 0.5)$, and $(0.8, 0.2)$ to these scenarios, respectively. We avoid setting  $(1, 0)$ for Performance First to prevent degenerate routing, always selecting the largest LLM. To enable utility-based optimization, following \citep{graphrouter}, we normalize performance and cost metrics to remove scale discrepancies. 


\paragraph{Baseline Methods.} To evaluate the effectiveness of our proposed TRouter, we compare it against two categories of baseline methods. The first category, Cold-Start Baselines, includes approaches that do not rely on in-domain training data $\mathcal{D}_\text{train}$: (1) Smallest LLM, always selecting the smallest available model to minimize inference cost; (2) Largest LLM, selecting the largest model to maximize performance; (3) Adaptive LLM, dynamically selecting more expensive model based on the user's tolerance for costs; and (4) Prompt LLM, utilizing an external LLM to select a candidate model based on a prompt that encodes the input query, candidate model pool, and user objectives. In this setting, TRouter is trained exclusively on synthetic data $\mathcal{D}_\text{syn}$ designed to simulate realistic task scenarios. 

The second category, In-Domain Baselines, includes methods that assume access to $\mathcal{D}_\text{train}$: (5) RouterDC~\citep{routerdc}, a dual-contrastive learning-based selector; (6) GraphRouter~\citep{graphrouter}, which models routing as edge prediction on a heterogeneous graph of tasks, queries, and models; (7) MetricRouter~\citep{omnirouter}, a regression-based method that predicts performance and cost using sentence embeddings; (8) FrugalGPT~\citep{frugalgpt}, which estimates generation quality under cost constraints; (9) C2MAB-V~\citep{C2MAB-V}, a contextual bandit approach for adaptive model selection; and (10) Oracle, a theoretical upper bound that selects the model with the highest utility using ground-truth cost and performance. In the in-domain setting, TRouter is also trained on $\mathcal{D}_{\text{train}}$ after annotating task type using LLMs. Additionally, we adapt RouterDC and MetricRouter to the cold-start setting using cross-dataset training for the cold-start comparisons.

\paragraph{Implementation Details.} We instantiate the data synthesis pipeline with gpt-4.1 and validate versatility by substituting gemini-2.5-flash. Performance is estimated via an LLM-as-a-judge using gpt-4.1-nano for cost efficiency (Table \ref{tab:mainllmasjudge}). Starting from six seed domains, we expand to ten (Table \ref{tab:domainnodegenprompt}); per domain, we generate up to ten subcategories and five difficulty levels (Tables \ref{tab:subcategorynodegenprompt}, \ref{tab:difficultydegenprompt}). The task type quality evaluator terminates after three consecutive no-change rounds (Table \ref{tab:tasktypequalityevaluator}). The QA pair generator produces 40 QA pairs per task profile (batch size 8). The resulting datasets comprise, for gpt-4.1: 10 domains, 103 subcategories, 447 difficulty nodes, and 17,880 QA pairs; and for gemini-2.5-flash: 10 domains, 98 subcategories, 380 difficulty nodes, and 15,200 QA pairs.

For the proposed TRouter, we operate at the difficulty level for $\mathcal{T}$, encoding queries and task-type descriptions with all-MiniLM-L6-v2~\footnote{\url{https://huggingface.co/sentence-transformers/all-MiniLM-L6-v2}} and mapping them into 256-dimensional vectors. Models are trained with learning rate 1e-4. We use 30 training and 10 validation QA pairs per task type in the cold-start setting. Due to page limitations, full Experimental Settings are provided in Appendix~\ref{app:experimalsetting}.
\subsection{Main Experimental Results}
\begin{table*}[t]                                                                           
\centering
\caption{Results across three user preference settings in both cold-start and in-domain scenarios. For in-domain training, candidate LLM performance is obtained using \textbf{traditional metrics}. Bold and underline indicate the best and second-best results under each user preference, respectively. We use ${\star}$ to denote routers trained on any three test tasks and evaluated on the remaining tasks (cross-domain). The symbols ${\blacktriangle}$ and ${\bullet}$ indicate that our data synthesis framework is derived using gpt-4.1 and gemini-2.5-flash, respectively.}
\begin{adjustbox}{max width=0.95\linewidth}
\begin{tabular}{ll
                *{3}{S[table-format=1.4]} S[table-format=1.4]
                *{3}{S[table-format=1.4]} 
                *{3}{S[table-format=1.4]} 
                S[table-format=1.4]}
\toprule
& & \multicolumn{4}{c}{Cost-first} & \multicolumn{3}{c}{Balance} & \multicolumn{3}{c}{Performance-first} & {} \\
\cmidrule(lr){3-6} \cmidrule(lr){7-9} \cmidrule(lr){10-12}
{Scenario} & {Method} 
& {Cost} & {Performance} & {Utility} & {}
& {Cost} & {Performance} & {Utility}
& {Cost} & {Performance} & {Utility}
& {Utility Sum} \\
\midrule
\multirow{7}{*}{Cold-start}
& Smallest LLM      & 0.0230 & 0.2004 & 0.0217 & & 0.0230 & 0.2004 & 0.0887 & 0.0230 & 0.2004 & 0.1557 & 0.2661 \\
& Largest LLM       & 0.3098 & 0.4383 & -0.1601 & & 0.3098 & 0.4383 & 0.0643 & 0.3098 & 0.4383 & 0.2887 & 0.1928 \\
& Adaptive LLM     & 0.0230  & 	0.2004  & 	0.0217  &   & 	0.0641  & 	0.4260  & 	0.1809  & 	0.3098	  & 0.4383	  & 0.2887	  & 0.4913\\

& Prompt LLM        & 0.0427 & 0.2131 & 0.0085 & & 0.1128 & 0.3996 & 0.1434 & 0.3098 & 0.4383 & 0.2887 & 0.4406 \\
& RouterDC$^{\star}$      & 0.0391 & 0.2548 & 0.0197 & & 0.0599 & 0.3580 & 0.1490 & 0.0650 & 0.3899 & 0.2989 & 0.4676 \\
& MetricRouter$^{\star}$  & 0.0325 & 0.1916 & 0.0123 & & 0.0446 & 0.2027 & 0.0703 & 0.1012 & 0.2118 & 0.1492 & 0.2319 \\
& Ours$^{\blacktriangle}$              & 0.0621 & 0.4257 & \textbf{0.0355} & & 0.0641 & 0.4262 & \textbf{0.1811} & 0.1721 & 0.4315 & \underline{0.3108} & \underline{0.5274} \\
& Ours$^{\bullet}$   &  0.0327	  & 0.3068	  & \underline{0.0352}	  &  & 0.0526   & 	0.4144  & 	 \underline{0.1809}	  & 0.0701	  & 0.4201	  & \textbf{0.3221}	  & \textbf{0.5382} \\
\midrule
\multirow{7}{*}{In-domain}
& RouterDC          & 0.0578 & 0.4367 & 0.0411 & & 0.0593 & 0.4290 & 0.1849 & 0.0660 & 0.4425 & 0.3408 & 0.5667 \\
& GraphRouter       & 0.0480 & 0.3059 & 0.0228 & & 0.0783 & 0.3976 & 0.1597 & 0.0628 & 0.4113 & 0.3165 & 0.4989 \\
& FrugalGPT         & 0.0230 & 0.2004 & 0.0217 & & 0.0890 & 0.4141 & 0.1625 & 0.3535 & 0.4267 & 0.2706 & 0.4548 \\
& C2MAB-V           & 0.0640 & 0.4213 & 0.0323 & & 0.0635 & 0.4127 & 0.1746 & 0.0758 & 0.4146 & 0.3166 & 0.5234 \\
& MetricRouter       & 0.0479 & 0.4125 & 0.0442 & & 0.0559 & 0.4382 & 0.1911 & 0.0957 & 0.4474 &  \underline{0.3388} & 0.5741 \\
& Ours$^{\blacktriangle}$              & 0.0393 & 0.4165 & \textbf{0.0518} & & 0.0477 & 0.4424 & \textbf{0.1974} & 0.0751 & 0.4479 & \textbf{0.3433} & \textbf{0.5925} \\
& Ours$^{\bullet}$   &  0.0400	 & 0.4127	 & \underline{0.0509}	 &  & 0.0481	 & 0.4353 & 	\underline{0.1936}  & 	0.0876  & 	0.4389	 & 0.3336 & 	 \underline{0.5836} \\
\midrule
\multicolumn{2}{c}{Oracle}  & 0.0354 & 0.4943 & 0.0705 & & 0.0417 & 0.5081 & 0.2332 & 0.0513 & 0.5127 & 0.3999 & 0.7037 \\
\bottomrule
\vspace{-10pt}
\end{tabular}
\end{adjustbox}
\label{tab:maintraditionmetric}
\end{table*}
\begin{table*}[t]
\centering
\caption{Results across three user preference settings in both cold-start and in-domain scenarios. For in-domain training, candidate LLM performance is obtained using \textbf{LLM-as-a-judge}. Bold and underline indicate the best and second-best results under each user preference, respectively. We use ${\star}$ to denote routers trained on any three test tasks and evaluated on the remaining tasks (cross-domain).}
\begin{adjustbox}{max width=0.95\linewidth}
\begin{tabular}{ll
                *{3}{S[table-format=1.4]} S[table-format=1.4]
                *{3}{S[table-format=1.4]} 
                *{3}{S[table-format=1.4]} 
                S[table-format=1.4]}
\toprule
& & \multicolumn{4}{c}{Cost-first} & \multicolumn{3}{c}{Balance} & \multicolumn{3}{c}{Performance-first} & {} \\
\cmidrule(lr){3-6} \cmidrule(lr){7-9} \cmidrule(lr){10-12}
{Scenario} & {Method} 
& {Cost} & {Performance} & {Utility} & {}
& {Cost} & {Performance} & {Utility}
& {Cost} & {Performance} & {Utility}
& {Utility Sum} \\
\midrule
\multirow{7}{*}{Cold-start}
& Smallest LLM     & 0.0226 & 0.3342 & 0.0488 & & 0.0226 & 0.3342 & 0.1558 & 0.0226 & 0.3342 & 0.2629 & 0.4675 \\
& Largest LLM      & 0.3082 & 0.5158 & -0.1434 & & 0.3082 & 0.5158 & 0.1038 & 0.3082 & 0.5158 & \underline{0.3510} & 0.3113 \\
& Adaptive LLM   & 0.0226 & 0.3342 & 0.0488 & & 0.0637 & 0.5024 & \underline{0.2194} & 0.3082 & 0.5158 & \underline{0.3510} & \underline{0.6191} \\
& Prompt LLM       & 0.0394 & 0.4053 & 0.0495 & & 0.1088 & 0.4853 & 0.1883 & 0.3082 & 0.5158 & \underline{0.3510} & 0.5888 \\
& RouterDC$^{\star}$         & 0.0713 & 0.5573 & \textbf{0.0544} & & 0.0914 & 0.5231 & 0.2159 & 0.1148 & 0.5297 & 0.2074 & 0.4777 \\
& MetricRouter$^{\star}$     & 0.0620 & 0.4967 & 0.0498 & & 0.1240 & 0.5397 & 0.2078 & 0.1833 & 0.5493 & 0.1830 & 0.4406 \\
& Ours             & 0.0617 & 0.4992 & \underline{0.0505} & & 0.0637 & 0.5030 & \textbf{0.2196} & 0.0740 & 0.4997 & \textbf{0.3850} & \textbf{0.6551} \\
\midrule
\multirow{6}{*}{In-domain}
& RouterDC         & 0.0369 & 0.4387 & \underline{0.0582} & & 0.0572 & 0.4577 & 0.2002 & 0.0638 & 0.4497 & 0.3470 & 0.6054 \\
& GraphRouter      & 0.0240 & 0.3395 & 0.0487 & & 0.0316 & 0.3616 & 0.1650 & 0.0337 & 0.3698 & 0.2891 & 0.5028 \\
& FrugalGPT        & 0.0230 & 0.3302 & 0.0476 & & 0.0899 & 0.3914 & 0.1508 & 0.1464 & 0.4872 & 0.3605 & 0.5589 \\
& C2MAB-V          & 0.0636 & 0.5003 & 0.0492 & & 0.0637 & 0.5024 & \underline{0.2194} & 0.1290 & 0.5054 & \underline{0.3785} & 0.6471 \\
& MetricRouter     & 0.0376 & 0.4526 & \textbf{0.0604} & & 0.0647 & 0.5024 & 0.2189 & 0.1106 & 0.4988 & 0.3769 & \underline{0.6512} \\
& Ours             & 0.0421 & 0.4705 & \textbf{0.0604} & & 0.0549 & 0.4982 & \textbf{0.2223} & 0.1080 & 0.5152 & \textbf{0.3906} & \textbf{0.6733} \\
\midrule
\multicolumn{2}{c}{Oracle}   & 0.0388 & 0.6390 & 0.0968 & & 0.0633 & 0.6984 & 0.3175 & 0.0690 & 0.7013 & 0.5472 & 0.9615 \\
\bottomrule
\end{tabular}
\end{adjustbox}
\label{tab:mainllmasjudge}
\vspace{-10pt}
\end{table*}
We evaluate our proposed task-profile guided data strategy and TRouter in the cold-start setting, where training data $\mathcal{D}_\text{train}$ from the test scenario is unavailable, and the in-domain setting, where such data can be collected. We report results under two evaluation protocols, including traditional metrics and LLM-as-a-judge in Table~\ref{tab:maintraditionmetric} and Table~\ref{tab:mainllmasjudge}. The results lead to three main conclusions.

First, the task-profile guided data strategy effectively mitigates the inability of LLM routers to adapt under cold-start settings. For example, TRouter trained on the synthetic dataset $\mathcal{D}_\text{syn}$ achieves substantial gains over the best baselines that do not rely on $\mathcal{D}_\text{train}$  (i.e., Adaptive LLM) across different evaluation protocols. When using LLM-as-a-judge in Table~\ref{tab:mainllmasjudge}, TRouter, trained solely on $\mathcal{D}_\text{syn}$, surpasses most in-domain baselines. However, methods relying on cross-domain training (RouterDC and MetricRouter) suffer from severe performance degradation due to the domain shift in the distribution of queries and corresponding metrics across different tasks, with the regression-based MetricRouter being particularly vulnerable. These findings indicate that LLM-driven data synthesis, conditioned only on domain descriptions, can generate effective supervision signals for routing, thus enabling cold-start routing in resource-constrained scenarios.

Second, our proposed TRouter demonstrates advantages in both cold-start and in-domain settings. For instance, in the in-domain scenario, TRouter consistently outperforms existing learning-based routing frameworks across diverse user preferences, whereas competing baselines (e.g., C2MAB-V) often exhibit large performance variance. This robustness possibly stems from TRouter’s task-type guided regularization, which injects structured task semantics beyond query semantics.

Third, although synthetic data is a strong choice for cold-start, TRouter still benefits from training on data drawn from the same distribution as the test scenario. As shown in Table~\ref{tab:maintraditionmetric}, the TRouter in the in-domain setting improves the utility sum by more than 0.05 relative to the cold-start counterpart, while this utility gap narrows considerably under LLM-as-a-judge in Table~\ref{tab:mainllmasjudge}. We suggest that this is due to differing performance distribution of candidate LLMs induced by different evaluation protocols, as corroborated by Fig.~\ref{fig:model_performance_disx} in the appendix. Consequently, for practical deployment, we recommend that end-users and router providers adopt aligned evaluation protocols, e.g., using LLM-as-a-judge consistently when deploying routers.

\subsection{Ablation Study}
\begin{figure}[t]
\centering
\includegraphics[width=\linewidth]{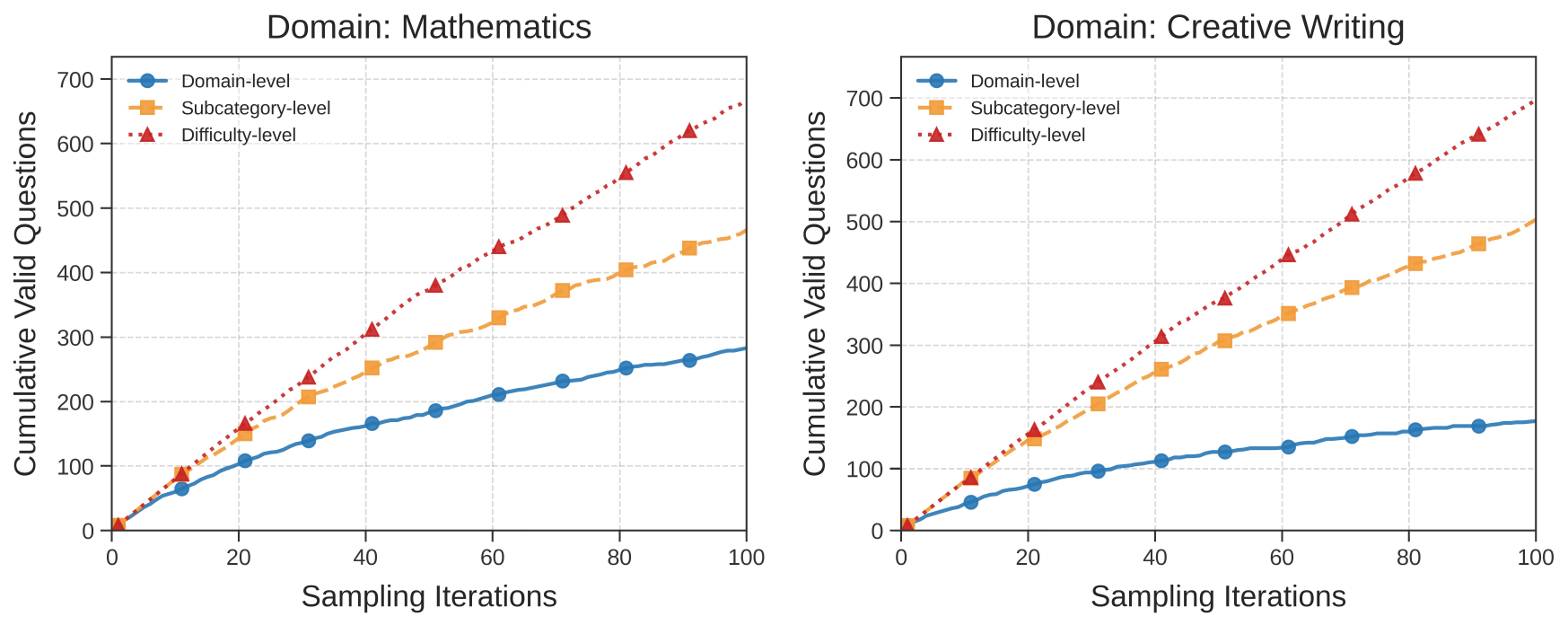}
\caption{Effect of using task types of different taxonomy level to construct task-profile on sampling efficiency on Mathematics and Creative Writing domains.}
\label{fig:abasamplingstrategies}
\end{figure}
\begin{figure}[t]
\centering
\includegraphics[width=\linewidth]{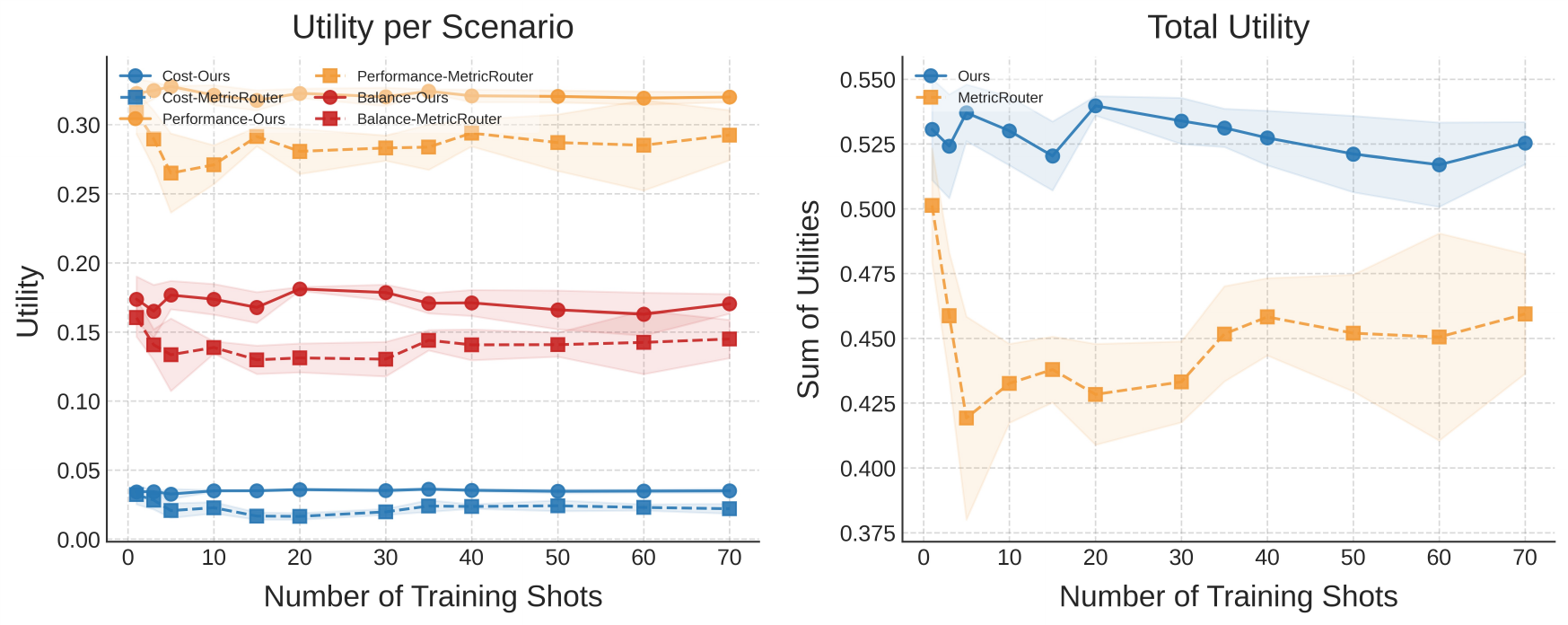}
\caption{Effect of shot  number  in the cold-start setting for TRouter and MetricRouter.}
\label{fig:abashotnumber}
\end{figure}
\begin{table}[t]
\centering
\caption{Effect of $\mathcal{T}$ across three taxonomy levels. Bold denotes the first best values under each user preferences.}
\label{tab:ablation_cold_in}
\begin{adjustbox}{max width=\linewidth}
\begin{tabular}{lcccc}
\toprule
Method & Cost & Balance & Performance & Utility Sum \\
\midrule
\multicolumn{5}{l}{\textit{Cold Start}} \\
Ours w/ domain regulation & 0.0343 & 0.1809 &\textbf{0.3250} & \textbf{0.5402} \\
Ours w/ subcategory regulation & 0.0350 & 0.1740 & 0.3204 & 0.5294 \\
Ours                      & \textbf{0.0355} & \textbf{0.1811} & 0.3108 & 0.5274 \\
\midrule
\multicolumn{5}{l}{\textit{In-domain}} \\
Ours w/ domain regulation       & 0.0505 & 0.1887 & 0.3249 & 0.5641 \\
Ours w/ subcategory regulation  & 0.0510 & 0.1920 & 0.3379 & 0.5809 \\
Ours                            & \textbf{0.0518} & \textbf{0.1974} & \textbf{0.3433} & \textbf{0.5925} \\
\bottomrule
\end{tabular}
\end{adjustbox}
\end{table}
\paragraph{Effect of task profiles across taxonomy levels.} We adopt a hierarchical task taxonomy, including domain, subcategory, and difficulty, in our data synthesis framework. In the question–answer pair generator, we instantiate a task profile using the difficulty-level task type as a conditional prompt to guide QA pair generation. To assess the impact of different taxonomy levels on sampling efficiency, we compare this approach with constructing task profiles at the domain and subcategory levels in Fig.~\ref{fig:abasamplingstrategies}. The results show that under a fixed number of sampling iterations, difficulty-conditioned profiles produce an almost linear increase in validated QA pairs, whereas domain and subcategory-conditioned profiles exhibit slowing gains due to accumulating redundancy across two representative domains. These results indicate that difficulty-level conditioning yields more diverse samples and better approximates the test-time query distribution in the cold-start routing setting.
\paragraph{Effect of shot number in cold-start setting.} We train TRouter on the synthetic dataset $\mathcal{D}_\text{syn}$ with varying shots per task in Fig. \ref{fig:abashotnumber}. Although our main experiments use 30 shots, performance with five or fewer shots is already near-optimal across diverse user preferences in LLM routing. This demonstrates that TRouter is sample-efficient and can be deployed cheaply in the cold-start setting.
\paragraph{Effect of task regularization in cold-start setting.} Removing the task regularization in Eq. \eqref{eq:elbo-r}, TRouter degrade to MetricRouter. As shown in Fig. \ref{fig:abashotnumber}, when trained on the same  $\mathcal{D}_\text{syn}$, TRouter consistently outperforms MetricRouter across all shot numbers. This advantage stems from incorporating a task-type prior, which enables TRouter to handle numerous task types and better cope with complex scenarios. We also observe that MetricRouter’s performance is unstable, which decreases with more shots initially but increases thereafter, while TRouter remains robust in increasing shot number, further validating its effectiveness.
\paragraph{Effect of $\mathcal{T}$ across taxonomy levels.} We instantiate TRouter with task types $\mathcal{T}$ drawn from the difficulty level of our taxonomy in the main experiments. In Table~\ref{tab:ablation_cold_in}, we compare alternatives that use domain and subcategory-level types. The results show that in in-domain setting, difficulty-based $\mathcal{T}$ yields the best performance, while domain-based $\mathcal{T}$  in cold-start settings, likely because finer-grained difficulty types introduce interference when task-type priors are not yet calibrated. Moreover, after in-domain adaptation, the difficulty-level  $\mathcal{T}$ benefits from its larger, more granular type space, which provides richer priors beyond query semantics, enabling more accurate routing.

\section{Conclusions}
We address the cold-start challenge in LLM routing by introducing a multi-level task-profile–guided data synthesis framework and TRouter, a task-type–aware LLM router. Our synthesis approach constructs diverse, taxonomy-aligned QA data that approximates test-time query distribution without requiring in-domain datasets. TRouter augments regression-based routing with latent task-type variables and a prior over the synthesized taxonomy, capturing structure beyond surface semantics. Across experiments on multiple LLM pools and evaluation protocols, we demonstrate our synthesis framework enables effective cold-start training, and TRouter consistently improves routing utility in both cold-start and in-domain settings. 
\section*{Limitations}
We discussed three imitations of our work. First, the task-profile–guided synthesis requires minimal human input to specify candidate domains (or a short application description from which the LLM can derive domains). This introduces unavoidable human intervention. In practical deployments (e.g., customer support), however, providers typically know the high-level context and can seed domains manually or automatically generate them using LLM. Compared with existing routers that generally require in-domain labeled data, our approach is better suited to cold-start settings. Moreover, allowing limited human input provides desirable control and flexibility, such as pruning low-quality child task types and injecting human priors to refine our three-level taxonomy to an application-specific taxonomy. Second, our synthesized QA pairs are not exhaustively validated. Prior work on data synthesis for training LLMs often relies on strict rule-based filtering to avoid performance collapse. In contrast, our setting models cost–performance relationships at the task-type level, i.e., across distributions of queries rather than individual QA pairs, making TRouter more robust to noise in any single QA pair and reducing the need for heavy-handed data quality control. The empirical results in Fig. \ref{fig:abashotnumber} also support this hypothesis. Third, using LLM-as-a-Judge may introduce model bias. In our main experiments, we mitigate this by selecting \texttt{gpt-4.1-nano-2025-04-14} as the judge model while evaluating candidate models from the Qwen and Gemma families, reducing the risk of self-preference bias. Notably, LLM-as-a-Judge aligns more closely with human preferences than traditional metrics. For instance, in tasks like GSM8K, conventional metrics may misjudge performance due to formatting issues, whereas LLM-based evaluation better captures semantic correctness. As discussed in Sec~ 4.2 (third point), consistency between the evaluation setup and the test scenario is crucial. In practice, we recommend aligning the evaluation protocol with the target use case and incorporating additional cross-checking mechanisms for robustness.

\section*{Acknowledgments}
We thank the UGC FITE (6460001), DOMG (9229193), and NTU Start-Up Grant
(\#023284-00001), Singapore.

\bibliography{custom}

\appendix
\section{Experimental Setting}
\label{app:experimalsetting}
\subsection{Datasets}
Following recent work~\citep{graphrouter}, we select data from four distinct tasks, with their statistics summarized in Table~\ref{tab:dataset}. The details of these four datasets are outlined as follows:

\begin{itemize}
    \item \textbf{Alpaca}~\citep{taori2023alpaca}: A 52k-sample hybrid question-answering dataset used for fine-tuning the Alpaca model. It is automatically constructed by prompting a language model to generate training instances based on a small set of human-written instructions.

    \item \textbf{GSM8K}~\citep{cobbe2021training}: A benchmark for evaluating multi-step mathematical reasoning, consisting of 8.5k linguistically diverse grade-school math word problems.

    \item \textbf{SQuAD}~\citep{rajpurkar2016squad}: A widely used reading comprehension benchmark comprising over 100k question–answer pairs derived from more than 500 Wikipedia articles.

    \item \textbf{Multi-News}~\citep{fabbri2019multi}: A multi-document summarization dataset containing 56k article–summary pairs. The articles are sourced from Newser.com, and the summaries are professionally written by editors.
\end{itemize}
\subsection{Baseliens}
\label{app:baselines}
To assess the effectiveness of our data synthesis strategy in the cold-start setting, as well as the performance of our proposed \textbf{TRouter}, we conduct extensive comparisons against two categories of baseline methods under distinct experimental scenarios.

\begin{itemize}
    \item \textbf{Smallest LLM}: Always selects the smallest available LLM, prioritizing cost efficiency over performance.
    \item \textbf{Largest LLM}: Always selects the largest available LLM, prioritizing performance regardless of cost.
    \item \textbf{Adaptive LLM}: Selects among the largest, medium, and smallest LLMs based on the user preference scenario. Specifically, it selects the largest LLM for the Performance First scenario, a medium-sized LLM for the Balanced scenario, and the smallest LLM for the Cost First scenario.
    \item \textbf{Prompt LLM}: Incorporates the query, candidate models, and objectives (e.g., prioritizing effectiveness) directly into the prompt, and feeds it into an external LLM (i.e.,  \texttt{gpt-4.1-2025-04-14} ) to select the most suitable LLM from a pool of candidates.
\end{itemize}
In this setting, TRouter is trained exclusively on synthetic data $\mathcal{D}_{\text{syn}}$, generated to simulate realistic training data of the test scenario.

When training data for the test task~$\mathcal{D}_{\text{train}}$ is available (i.e., the in-domain setting), TRouter is evaluated against six strong baselines:

\begin{itemize}
    \item \textbf{RouterDC}~\citep{routerdc}: introduces a query-based routing method trained with dual contrastive learning losses, where an encoder and LLM embeddings are jointly optimized to effectively distinguish between capable models, even when multiple LLMs perform well on a given query.
    \item \textbf{GraphRouter}~\citep{graphrouter}: models the LLM selection process as an edge prediction task on a heterogeneous graph of tasks, queries, and LLMs, leveraging a heterogeneous GNN to capture contextual interactions and generalize to new LLMs without retraining.
    \item \textbf{FrugalGPT}~\citep{frugalgpt}: utilizes a pre-trained language model to predict the score of the
generation result of all LLMs given a query, and then selects the LLM with the highest score within
a given cost. We use DistilBert~\citep{sanh2019distilbert} as the router’s backbone model.
    \item \textbf{C2MAB-V}~\citep{C2MAB-V}: uses a bandit-based model for LLM selection, which regards each
LLM as an arm that implements an exploration mechanism to find a better solution.
    \item \textbf{MetricRouter}~\citep{omnirouter}:  integrate multiple multilayer perceptron (MLP) layers with sentence transformer embeddings (specifically, the all-MiniLM-L6-v2 model\footnote{\url{https://huggingface.co/sentence-transformers/all-MiniLM-L6-v2}}) to develop regression models that estimate both the computational cost and performance of various large language models (LLMs) in response to a given input query.
    \item \textbf{Oracle}: A theoretical upper-bound baseline that assumes access to the ground-truth performance and cost for each query, and always selects the model yielding the highest utility.
\end{itemize}
In this setting, TRouter is trained on $\mathcal{D}_\text{train}$, which is realistic training data of the target test scenario. Additionally, each query in $\mathcal{D}_\text{train}$ is annotated with the task type at difficulty level.

To further evaluate the generalization ability of existing LLM routers under distribution shift, we assess RouterDC and MetricRouter in a cross-dataset cold-start setting. Specifically, we train on any three of the datasets listed in Table~\ref{tab:modelstat} and evaluate on the remaining one. We report the average performance, cost, and utility across all four datasets.

It is essential to note that RouterDC and GraphRouter are classification-based approaches that directly predict the optimal LLM. In contrast, MetricRouter employs a regression-based strategy to estimate both performance and cost, subsequently maximizing utility. In this context, our TRouter can be viewed as a regularized variant of MetricRouter, where the regularization term introduced in Eq.~\eqref{eq:elbo-r} is designed to enhance generalization, particularly under cold-start conditions. 

We exclude OmniRouter~\citep{omnirouter} and Carrot~\citep{carrot} from our comparison, as both adopt the same regression-based paradigm as MetricRouter, predicting performance and cost, then selecting models via utility maximization. However, OmniRouter is tailored for the AIOS setting, which schedules large models to maximize utility over a batch of queries, diverging from our focus on query-level routing. HybridLLM~\citep{hybridllm} is similarly excluded, as it is designed for two-model scenarios and is not directly applicable to our multi-model framework.

\subsection{Implementation Details}
\label{app:imple}
\paragraph{LLM Usage.} For the candidate LLMs used for routing, we primarily select the Qwen 3 series in our main experiments. In addition, we conduct supplementary experiments on several closed-source models, including the Gemini series and the Doubao-Seed models. For Qwen 3 models, we access them via the DashScope API provided by Alibaba Cloud\footnote{\url{https://bailian.console.aliyun.com/}}. For the other models, we utilize third-party API providers to obtain model outputs. 

For our task-profle guided data synthesis framework, we primarily utilize the \texttt{gpt-4.1-2025-04-14} model to instantiate key components, including the Task Type Generator, Task Type Quality Evaluator, and Question-Answer Pair Generator. To validate the generalizability of our approach, we further conduct experiments using the \texttt{gemini-2.5-flash} model as a substitute engine for all components of our synthesis framework.
\paragraph{Cost Analysis for Data Synthesis} 
The pricing for \texttt{gpt-4.1-2025-04-14} is \$2 per million input tokens and \$8 per million output tokens, whereas \texttt{gemini-2.5-flash} costs \$0.30 and \$2.50 per million input and output tokens, respectively. During QA generation, \texttt{gpt-4.1-2025-04-14} processed approximately 0.91M input tokens and generated 1.56M output tokens, resulting in a cost of \$14.34. In contrast, \texttt{gemini-2.5-flash} consumed 0.78M input tokens and produced 3.92M output tokens, with a total cost of \$10.03. The combined cost of data synthesis was approximately \$24.37. 
\paragraph{LLM as a Judge for Performance Estimating.} Beyond traditional metrics (e.g., token- or subword-level F1 for summarization), we estimate performance using an LLM-as-a-judge~\citep{gu2024survey} given the question, the ground-truth answer, and the answer produced by the candidate LLM. This choice is motivated by two considerations. First, prior work typically assumes that the evaluation protocol of the test scenario is known in advance and then derives router training data by applying that protocol to estimate performance. In realistic deployments, especially in the cold-start setting, this assumption may not hold, and a fixed evaluation protocol may be unavailable or misaligned with downstream needs. Second, traditional surface-form metrics exhibit well-documented limitations. For example, F1 implicitly equates lexical overlap with semantic equivalence, whereas summary quality depends on higher-level properties such as content selection, factual consistency, discourse structure, and readability. Our LLM-as-a-judge prompt is provided in Table \ref{tab:evaluator-prompt}, and we use the \texttt{gpt-4.1-nano-2025-04-14} model as the judging LLM to reduce costs. 
\paragraph{Details of Task Profile guided Data Synthesis.}
To construct a task profile-guided synthetic dataset, we leverage \texttt{gpt-4.1-2025-04-14} as an LLM engine to instantiate the three key components, including the task type generator, the task type quality evaluator, and the question-answer pair generator. Additional experiments are conducted using \texttt{gemini-2.5-flash} to validate the versatility of our framework. 

Since our evaluation spans a wide range of tasks, we begin by aggregating six representative domains from established LLM benchmark\footnote{\url{https://lmarena.ai/leaderboard/text}}, involving mathematics, creative writing, commonsense knowledge, programming, long-context understanding, and reading comprehension. These domains serve as seed categories that are expanded to ten broader domains via using the prompt in Table~\ref{tab:domainnodegenprompt} to derive the LLM engine, reducing manual intervention in domain specification.

For each domain, \textbf{Task Type Generator} produces up to ten subcategory nodes, and each subcategory is further expanded into no more than five difficulty levels, forming a three-level task taxonomy. The corresponding prompts for the subcategory level and difficulty level are shown in the Table. \ref{tab:subcategorynodegenprompt} and Table. \ref{tab:difficultydegenprompt}.

\textbf{Task Type Quality Evaluator} iteratively refines the candidate child task-type set and terminates when no modifications are warranted across three consecutive iterations, and the current set is adopted. The prompts used for task-set quality assessment and refinement are presented in Table \ref{tab:tasktypequalityevaluator}.

\textbf{Question-Answer Pair Generator}, we configure it to produce questions in batches of eight, generating a total of 40 QA pairs per task profile. We set the semantic similarity threshold to 0.9 in order to remove literally identical or near-duplicate examples. As shown in Appendix~\ref{appendix:quaqualityofqapairs}, our diversity analysis confirms that the generated data spans a broad range of topics, primarily due to our multi-level, task-type-guided generation framework. Thus, simple semantic filtering suffices for our objective. Users seeking further refinement may employ more advanced LLM-based filtering methods, though at a significantly higher computational cost. To estimate performance with minimal annotation cost for router training, we employ \texttt{gpt-4.1-nano} as a lightweight annotator.

The resulting synthetic dataset $\mathcal{D}_{\text{sys}}$ includes: (i) for \texttt{gpt-4.1-2025-04-14}, 10 domains, 103 subcategory nodes, 447 difficulty nodes, and 17,880 QA pairs; (ii) for  \texttt{gemini-2.5-flash}, the same number of domains but with 98 subcategory nodes, 380 difficulty nodes and 15200 QA pairs. 

\paragraph{Detailed of Task-type Guided Routing Framework.} In our proposed task-type aware router framework (TRouter), we use the task types at the difficulty level to instantiate $\mathcal{T}$. We utilize all-MiniLM-L6-v2 as the text encoder and map the query and description embedding to a 256-dimensional latent space. We set $\tau$ as 0.07. In the cold-start setting, we sample 30 QA pairs per task type for training and 10 for validation. All models are trained using the Adam optimizer with a learning rate of 1e-4, consistent across main experiments. For each model-metric pair, we utilize a two-layer MLP as the regressor to predict the performance and cost.
\section{Supplementary Theoretical Analysis}
\label{app:ssupptheoana}
In this section, we provide the proof of Eq.~\eqref{eq:elbo-r}. We start from the marginal likelihood of the observed triplet $(q,h,m)$ as follows:
\begin{equation}
\small
\log p_\theta(h \mid q,m)
= \log\!\sum_{t \in \mathcal{T}} p_\theta(h,t \mid q,m).
\end{equation}
After introduce the variational distribution \(q_{\phi}(t\mid q)\) and we rewrite $\log p_\theta(h \mid q,m)$ to:
\begin{small}
\begin{equation}
\begin{aligned}
\log p_\theta(h \mid q,m)
&= \log\!\sum_{t} q_{\phi}(t\mid q)\,
      \frac{p_\theta(h,t \mid q,m)}{q_{\phi}(t\mid q)} \notag\\
&= \log \mathbb{E}_{q_{\phi}(t\mid q)}
      \!\left[\frac{p_\theta(h,t \mid q,m)}{q_{\phi}(t\mid q)}\right]\!.
\end{aligned}
\end{equation}
\end{small}
\noindent By applying Jensen’s inequality and because \(\log\) is concave, we obtain 
\begin{equation}
\small
\log p_\theta(h \mid q,m)
\;\ge\;
\mathbb{E}_{q_{\phi}(t\mid q)}
\Bigl[\log p_\theta(h,t \mid q,m) - \log q_{\phi}(t\mid q)\Bigr].
\label{eq:jensen}
\end{equation}
Using the conditional independence assumption
\(p_\theta(h,t \mid q,m)=p_\theta(h\mid t,m)\,p(t\mid q)\),
Eq.~\eqref{eq:jensen} becomes
\begin{small}
\begin{equation}
\begin{aligned}
\log p_\theta(h \mid q,m)
\ge\;
&\mathbb{E}_{q_{\phi}(t\mid q)}\!\left[\log p_\theta(h \mid t,m)\right] \notag\\
&\quad-\mathbb{E}_{q_{\phi}(t\mid q)}
      \!\left[\log\frac{q_{\phi}(t\mid q)}{p(t\mid q)}\right]\!.
\end{aligned}
\end{equation}
\end{small}
The second expectation is the Kullback–Leibler divergence. Hence, we further obtain:
\begin{equation}
\small
\boxed{
\begin{aligned}
\log p_\theta(h \mid q,m) &\;\ge\;
\mathbb{E}_{q_{\phi}(t\mid q)}\!\left[\log p_\theta(h \mid t,m)\right]\\
&-
\mathrm{KL}\!\left(q_{\phi}(t\mid q)\,\|\,p(t\mid q)\right)
\end{aligned}
}
\end{equation}
Finally, we prove Eq.~\eqref{eq:elbo-r}.

\section{Supplementary Experiments}
\begin{table}[!t]
\centering
\caption{Overview of Supplementary Datasets.}
\label{tab:suppdataset}
\begin{adjustbox}{width=\linewidth}
\begin{tabular}{l l l r}
\toprule
\textbf{Dataset} & \textbf{Task Type} & \textbf{Metric} & \textbf{Cases} \\
\midrule
WMT~\citep{wmt}      & Translation            & BLEU       & 1000 \\
MBPP~\citep{mbpp}       & Programming           & Pass@1  & 964 \\
LegalBench~\citep{guha2023legalbench}       & Legal Domain   & F1        & 1000 \\
MedMcQA~\citep{medmcqa}  & Medical QA              & F1        & 1000 \\
\bottomrule
\end{tabular}
\end{adjustbox}
\vspace{-5pt}
\end{table}
\label{app:suppleexp}
\begin{table}[t]
\centering
\caption{Ablation study of different designs of TRouter.}
\label{tab:suppablation_component}
\begin{adjustbox}{max width=\linewidth}
\begin{tabular}{lcccc}
\toprule
Method & Cost & Balance & Performance & Utility Sum \\
\midrule
\multicolumn{5}{l}{\textit{Cold Start}} \\
AgenticRouter             & 0.0393 & 0.1771 & 0.3161 & 0.5325 \\
Ours w/o  Assumptions (1)  & 0.0357 & 0.1602 &0.3200 & 0.5159 \\
Ours                      & 0.0355 & 0.1811 & 0.3108 & 0.5274 \\
\midrule
\multicolumn{5}{l}{\textit{In-domain}} \\
Ours w/o  Assumptions (1)       & 0.0517  & 0.1964 & 0.3445 &0.5926 \\
Ours                            & 0.0518 & 0.1974 & 0.3433 & 0.5925 \\
\bottomrule
\end{tabular}
\end{adjustbox}
\end{table}

\begin{table*}[ht]
\centering
\caption{Results across three user preference settings in both cold-start and in-domain scenarios. For in-domain training, candidate LLM performance is obtained using \textbf{traditional metrics}. Bold indicate the best results under each user preference, respectively. We use commercial closed-source models in Table~\ref{tab:modelstat}.}
\label{tab:geminigroup}
\begin{adjustbox}{max width=\textwidth}
\begin{tabular}{ll
                *{3}{S[table-format=1.4]}
                *{3}{S[table-format=1.4]}
                *{3}{S[table-format=1.4]}
                S[table-format=1.4]}
\toprule
& & \multicolumn{3}{c}{Cost Preference} & \multicolumn{3}{c}{Balance Preference} & \multicolumn{3}{c}{Performance Preference} & {Utility Sum} \\
\cmidrule(lr){3-5} \cmidrule(lr){6-8} \cmidrule(lr){9-11}
Scenario & Method &
{Cost} & {Performance} & {Utility} &
{Cost} & {Performance} & {Utility} &
{Cost} & {Performance} & {Utility} &
{} \\
\midrule
Cold-start & Smallest LLM   & 0.0182 & 0.3987 & 0.0652 & 0.0182 & 0.3987 & 0.1903 & 0.0182 & 0.3987 & 0.3153 & 0.5708 \\
           & Largest LLM    & 0.4473 & 0.4290 & -0.2721 & 0.4473 & 0.4290 & -0.0092 & 0.4473 & 0.4290 & 0.2537 & -0.0275 \\
           & Adaptive LLM & 0.0182 & 0.3987 & 0.0652 & 0.0157 & 0.4224 & 0.2034 & 0.4473 & 0.4290 & 0.2537 & 0.5223 \\
           & Prompt LLM     & 0.0182 & 0.3987 & 0.0652 & 0.0194 & 0.4099 & 0.1953 & 0.4473 & 0.4290 & 0.2537 & 0.5142 \\
           & Ours           & 0.0165 & 0.4182 & \textbf{0.0704} & 0.0169 & 0.4113 & \textbf{0.1972} & 0.0183 & 0.4044 & \textbf{0.3199} & \textbf{0.5875} \\
\midrule
In-domain  & RouterDC       & 0.0154 & 0.4258 & 0.0729 & 0.0186 & 0.4291 & 0.2052 & 0.0169 & 0.4301 & 0.3407 & 0.6188 \\
           & GraphRouter    & 0.0230 & 0.4197 & 0.0655 & 0.0271 & 0.4086 & 0.1908 & 0.0184 & 0.4174 & 0.3302 & 0.5865 \\
           & FrugalGPT      & 0.0259 & 0.4360 & 0.0665 & 0.0509 & 0.4289 & 0.1890 & 0.0528 & 0.4286 & 0.3323 & 0.5878 \\
           & C2MAB-V        & 0.0163 & 0.4124 & 0.0694 & 0.0159 & 0.4153 & 0.1997 & 0.1464 & 0.4471 & 0.3284 & 0.5975 \\
           & MetricRouter   & 0.0137 & 0.4233 & 0.0737 & 0.0167 & 0.4245 & 0.2039 & 0.0231 & 0.4372 & 0.3451 & 0.6227 \\
           & Ours           & 0.0127 & 0.4311 & \textbf{0.0760} & 0.0140 & 0.4354 & \textbf{0.2107} & 0.0146 & 0.4362 & \textbf{0.3460} & \textbf{0.6327} \\
\midrule
Oracle     &                & 0.0117 & 0.4891 & 0.0884 & 0.0161 & 0.4975 & 0.2407 & 0.0197 & 0.4994 & 0.3956 & 0.7247 \\
\bottomrule
\end{tabular}
\end{adjustbox}
\end{table*}

\begin{table*}[ht]
\centering
\caption{Results across three user preference settings in both cold-start and in-domain scenarios on four supplementary datasets. For in-domain training, candidate LLM performance is obtained using \textbf{traditional metrics}.}
\label{tab:tablesupp}
\begin{adjustbox}{max width=\textwidth}
\begin{tabular}{ll
                *{3}{S[table-format=1.4]}
                *{3}{S[table-format=1.4]}
                *{3}{S[table-format=1.4]}
                S[table-format=1.4]}
\toprule
& & \multicolumn{3}{c}{Cost Preference} 
  & \multicolumn{3}{c}{Balance Preference} 
  & \multicolumn{3}{c}{Performance Preference} 
  & {Utility Sum} \\
\cmidrule(lr){3-5} 
\cmidrule(lr){6-8} 
\cmidrule(lr){9-11}

Scenario & Method &
{Cost} & {Performance} & {Utility} &
{Cost} & {Performance} & {Utility} &
{Cost} & {Performance} & {Utility} &
{} \\
\midrule

Cold-start 
& Smallest LLM 
& 0.0117 & 0.2977 & 0.0052
& 0.0117 & 0.2977 & 0.1430
& 0.0117 & 0.2977 & 0.2358
& 0.4290 \\

& Largest LLM 
& 0.1269 & 0.3857 & -0.0244
& 0.1269 & 0.3857 & 0.1294
& 0.1269 & 0.3857 & 0.2832
& 0.3882 \\

& Adaptive LLM
& 0.0117 & 0.2977 & 0.0052
& 0.0384 & 0.3496 & 0.1556
& 0.1269 & 0.3857 & 0.2832
& 0.4890 \\

& Ours
& 0.0125 & 0.3211 & 0.0543
& 0.0360 & 0.3490 & 0.1563
& 0.0467 & 0.3642 & 0.2820
& 0.4926 \\

\midrule

In-domain
& RouterDC
& 0.0258 & 0.3426 & 0.0479
& 0.0238 & 0.3430 & 0.1596
& 0.0374 & 0.3519 & 0.2740
& 0.4815 \\

& MetricRouter
& 0.0147 & 0.3324 & 0.0547
& 0.0563 & 0.3734 & 0.1586
& 0.0457 & 0.3716 & 0.2881
& 0.5014 \\

& Ours
& 0.0129 & 0.3242 & 0.0545
& 0.0416 & 0.3757 & 0.1671
& 0.0504 & 0.3766 & 0.2912
& 0.5128 \\

& ORACL
& 0.0151 & 0.4080 & 0.0695
& 0.0199 & 0.4172 & 0.1987
& 0.0289 & 0.4223 & 0.3321
& 0.6002 \\
\bottomrule
\end{tabular}
\end{adjustbox}
\end{table*}

\begin{figure}[t]
\centering
\includegraphics[width=\linewidth]{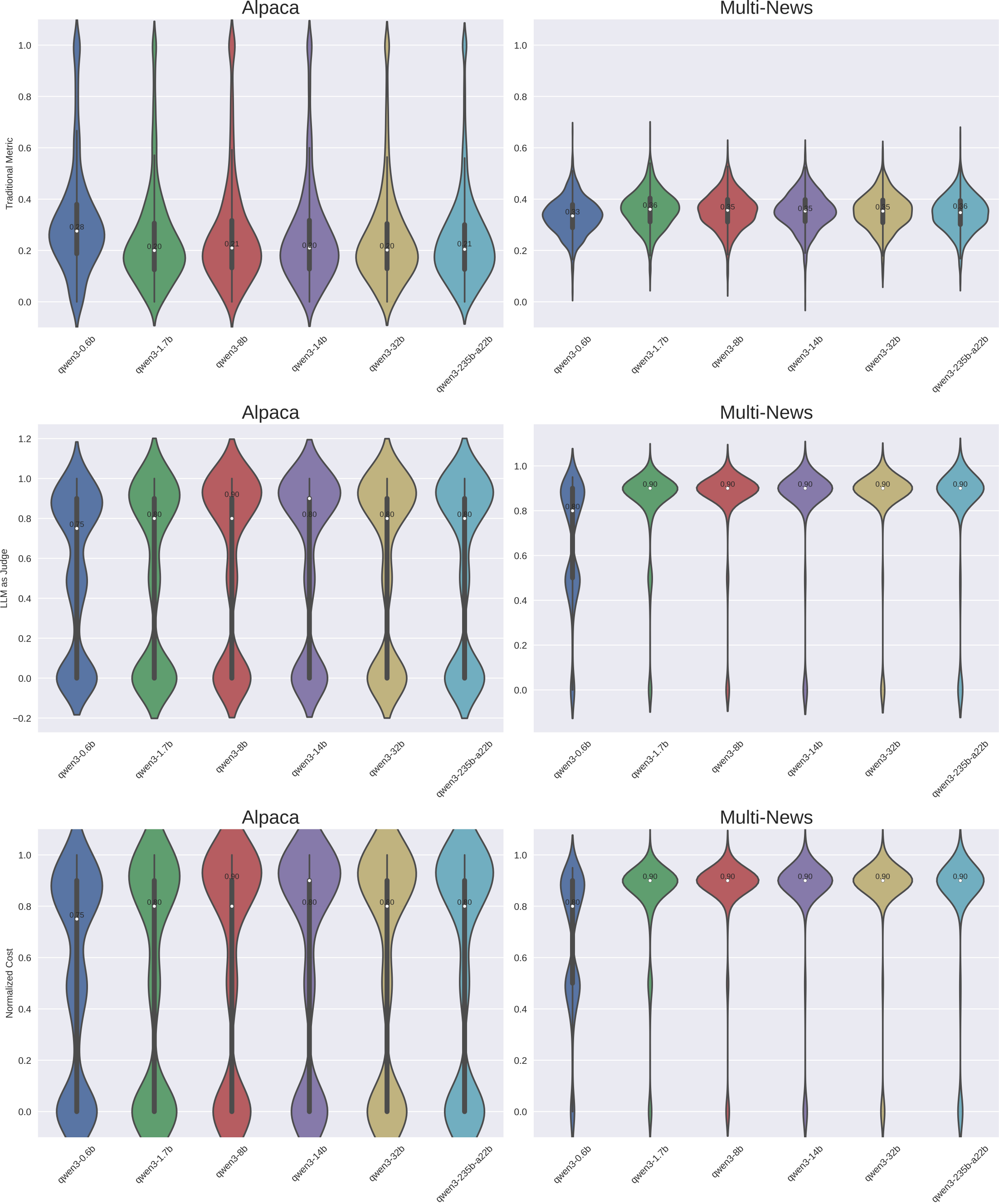}
\caption{The normalized cost and performance distributions for six Qwen-series models on Alpaca and Multi-News datasets, evaluated using traditional metrics. }
\label{fig:model_performance_disx}
\end{figure}
\begin{figure}[t]
\centering
\includegraphics[width=0.8\linewidth]{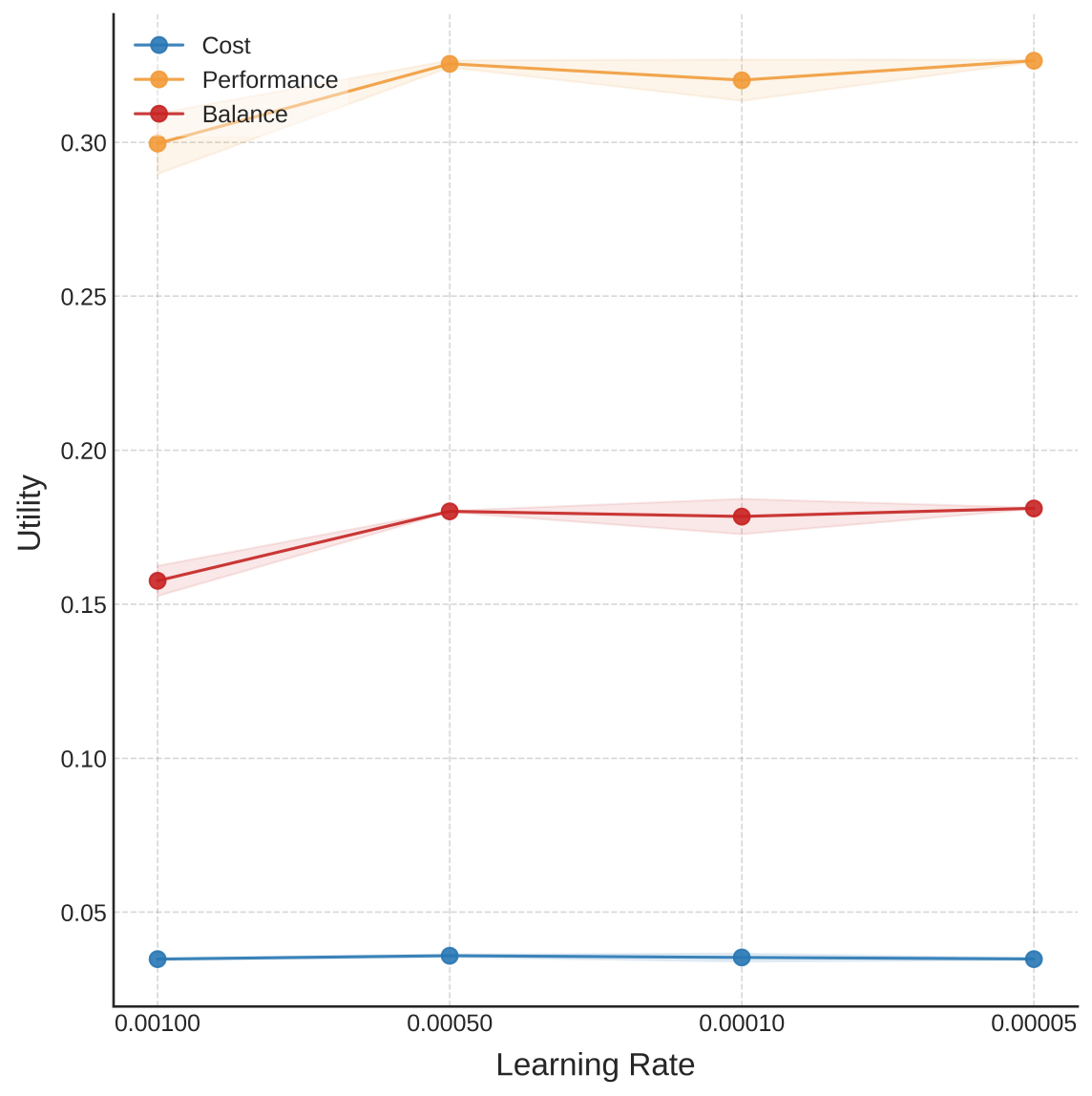}
\caption{Ablation on learning rate for TRouter in the cold-start setting. For each learning rate setting, results are averaged over five independent runs.}
\label{fig:lrrate}
\end{figure}
\begin{figure*}[t]
  \centering

  \begin{subfigure}[b]{0.33\linewidth}
    \centering
    \includegraphics[width=\linewidth]{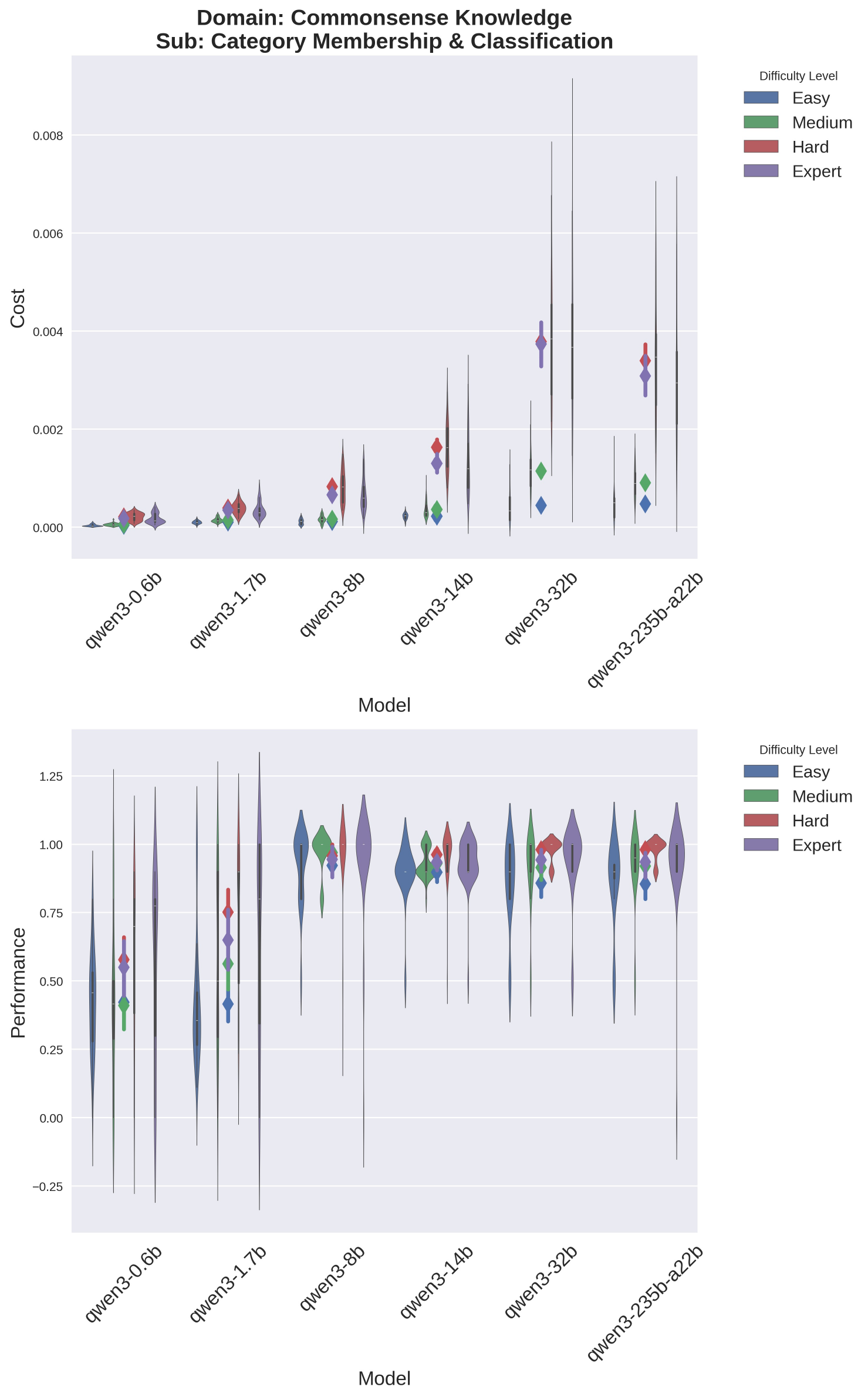}
  \end{subfigure}\hfill
  \begin{subfigure}[b]{0.33\linewidth}
    \centering
    \includegraphics[width=\linewidth]{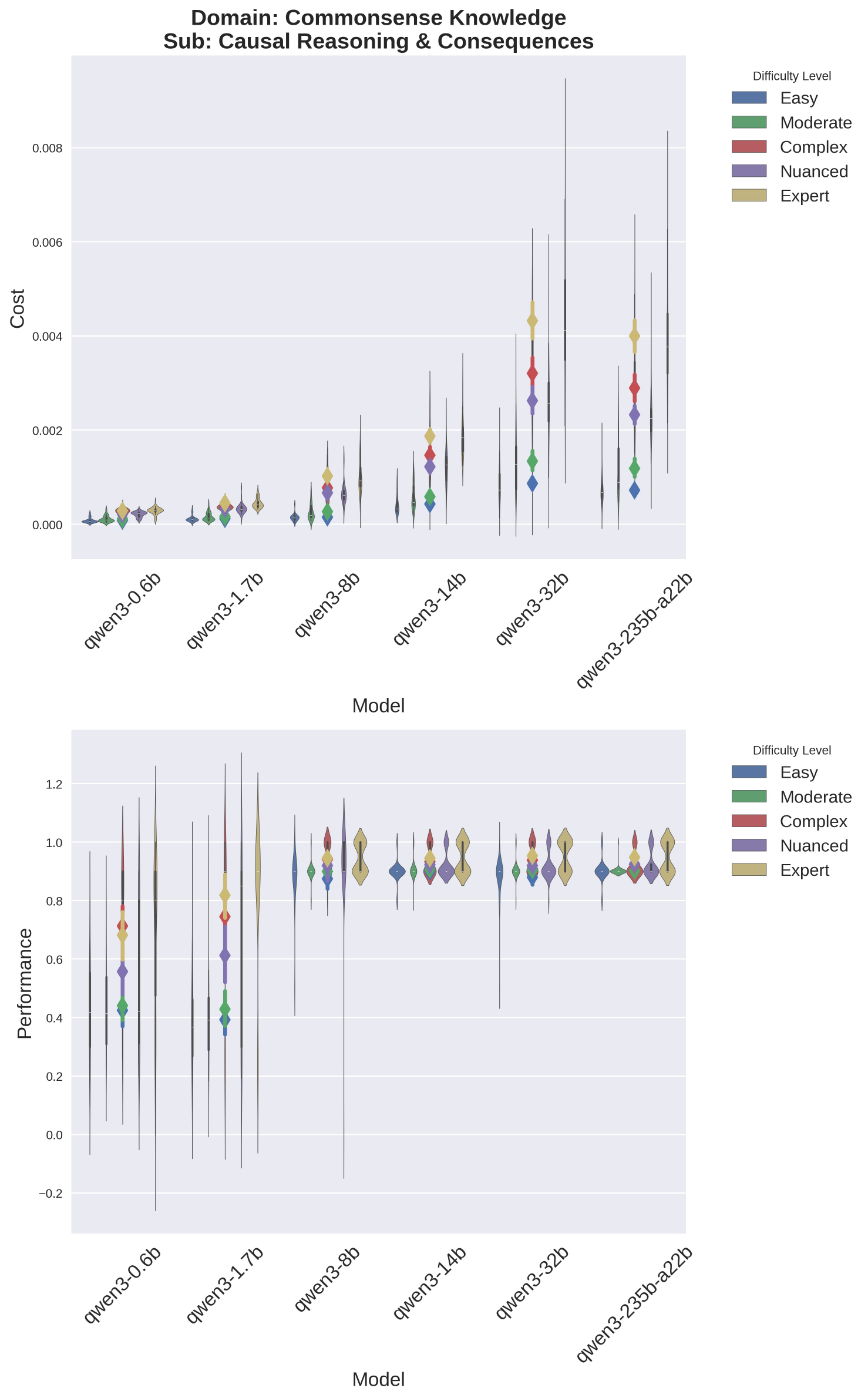}
  \end{subfigure}\hfill
  \begin{subfigure}[b]{0.33\linewidth}
    \centering
    \includegraphics[width=\linewidth]{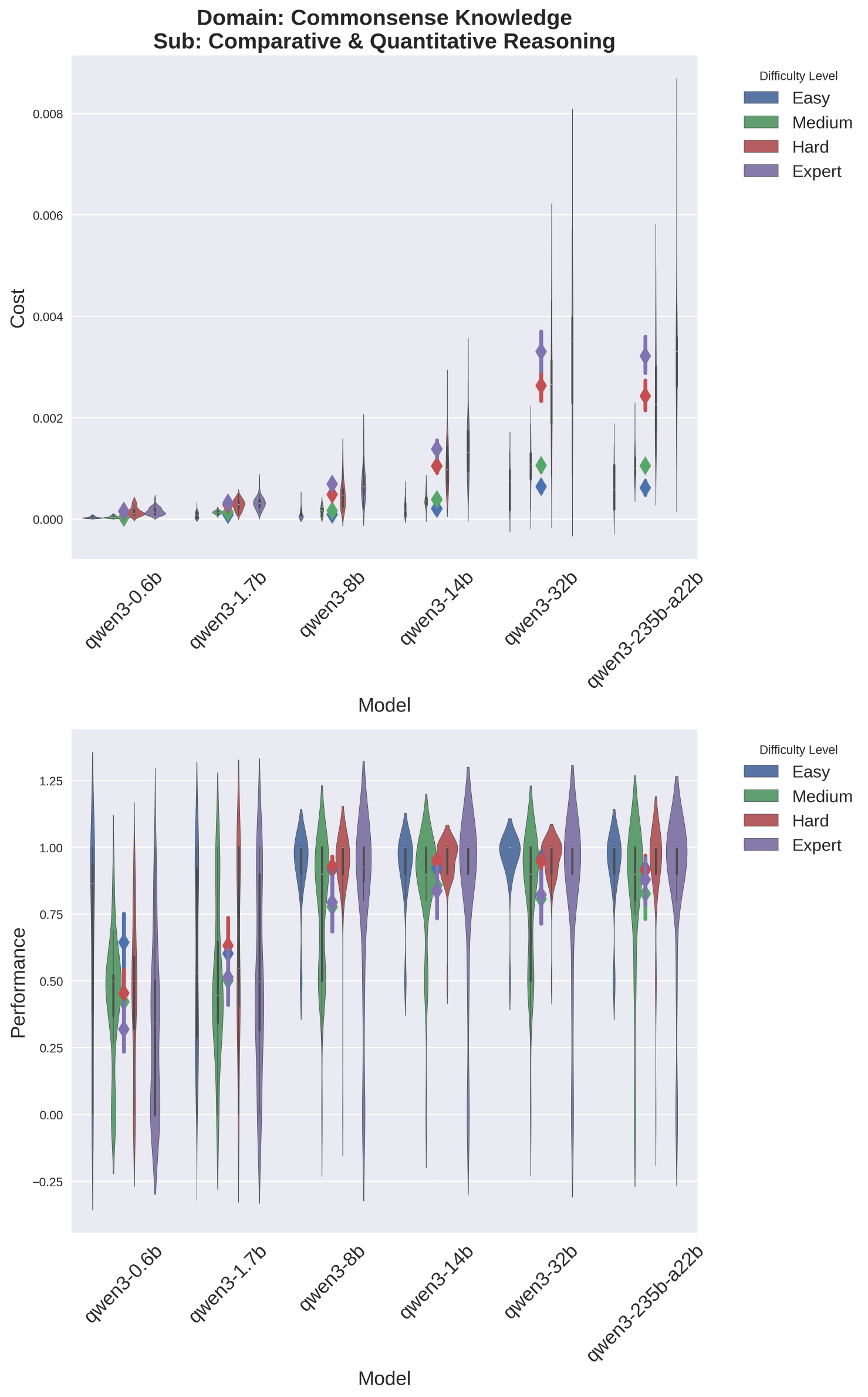}
  \end{subfigure}
  \caption{Visualization of cost and performance distributions for six Qwen-series models across representative subcategories of varying difficulty in the Commonsense Knowledge domain, derived from the synthesized dataset $\mathcal{D}_{\text{syn}}$.}
  \label{fig:grid3_commonsese}
\end{figure*}

\begin{figure*}[t]
  \centering
  \vspace{0.6em}
  \begin{subfigure}[b]{0.33\linewidth}
    \centering
    \includegraphics[width=\linewidth]{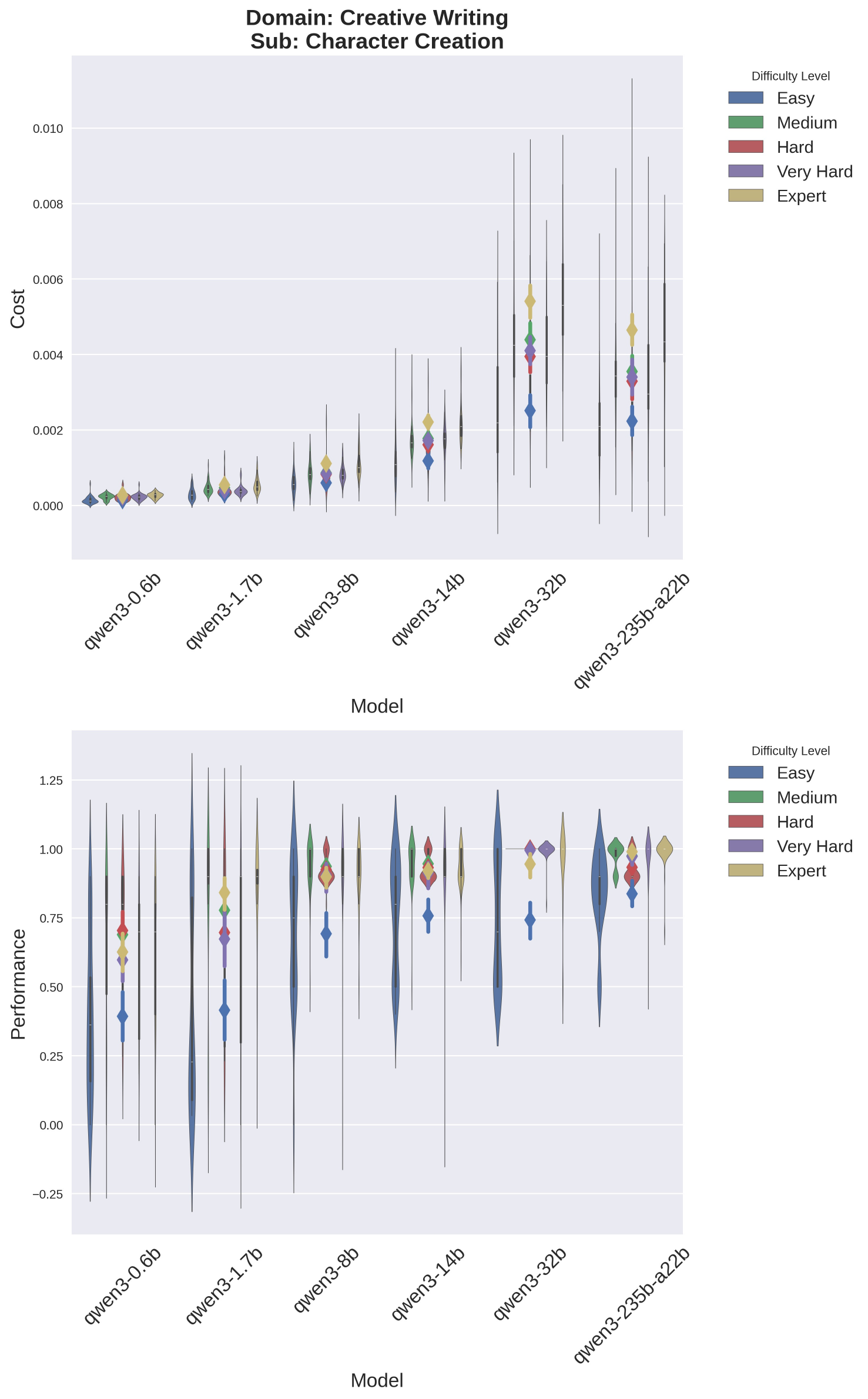}
  \end{subfigure}\hfill
  \begin{subfigure}[b]{0.33\linewidth}
    \centering
    \includegraphics[width=\linewidth]{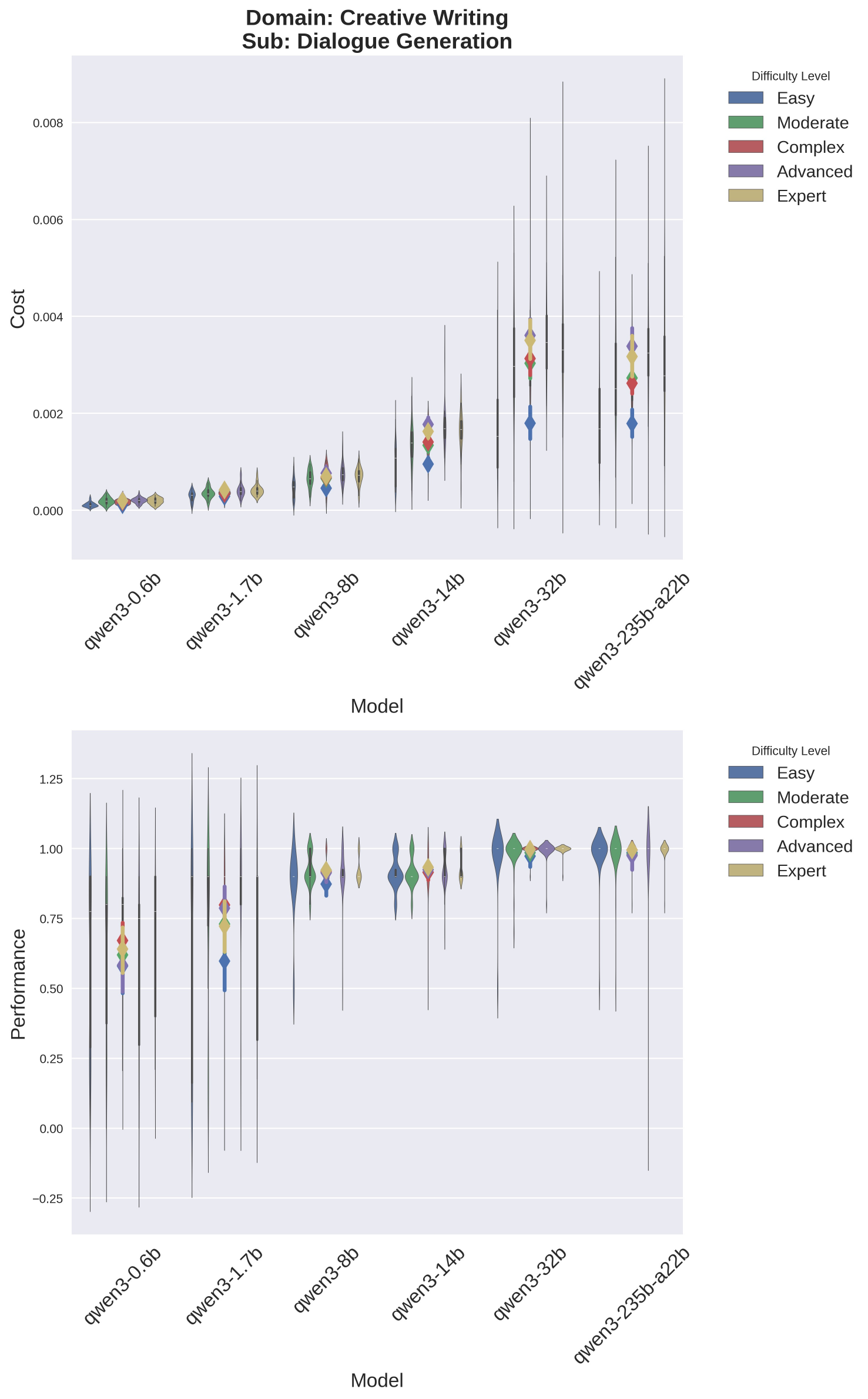}
  \end{subfigure}\hfill
  \begin{subfigure}[b]{0.33\linewidth}
    \centering
    \includegraphics[width=\linewidth]{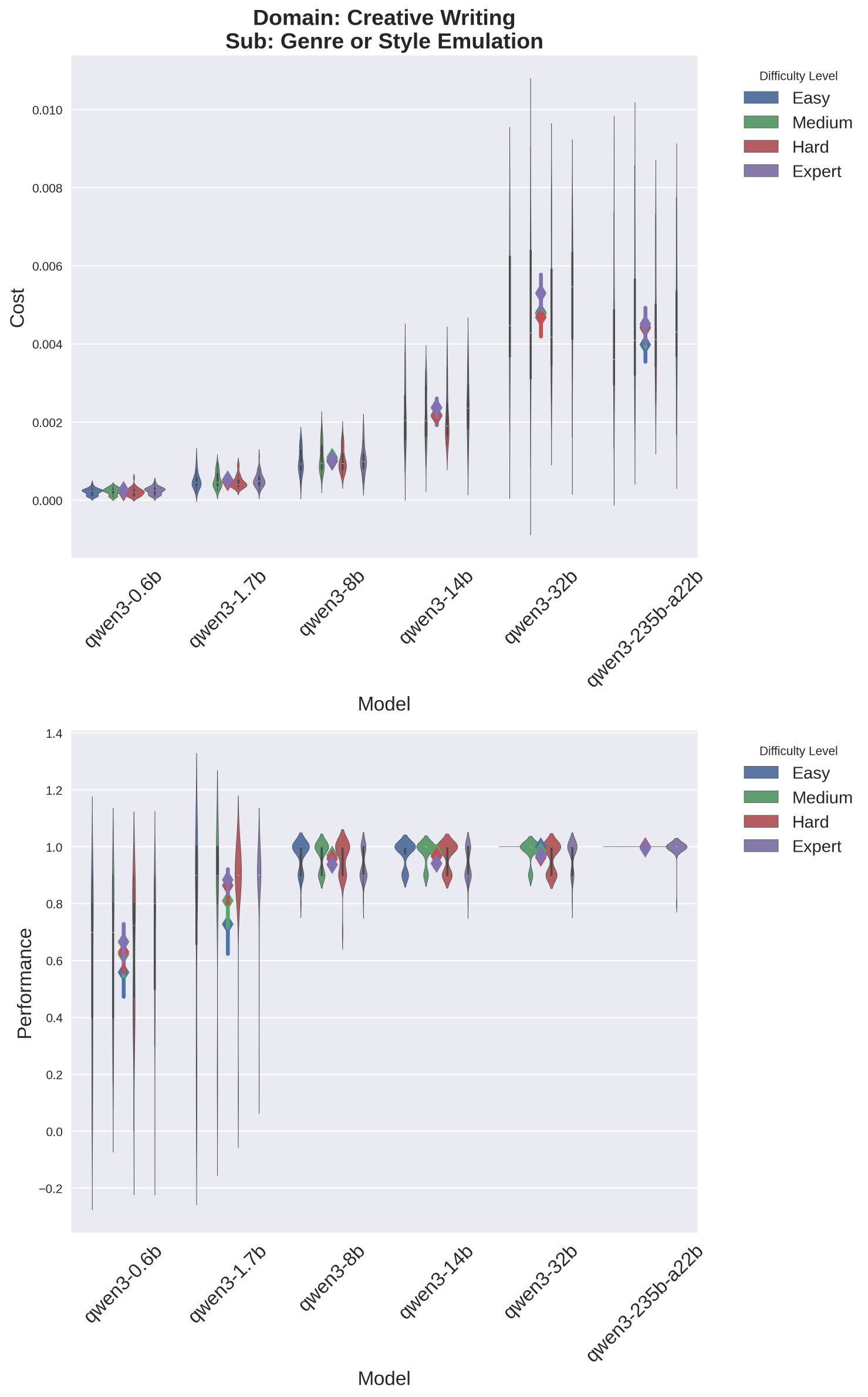}
  \end{subfigure}

  \caption{Visualization of cost and performance distributions for six Qwen-series models across representative subcategories of varying difficulty in the Creative Writing domain, derived from the synthesized dataset $\mathcal{D}_{\text{syn}}$.}
  \label{fig:grid9creative}
\end{figure*}

\begin{figure*}[t]
  \centering
  \vspace{0.6em}
  \begin{subfigure}[b]{0.33\linewidth}
    \centering
    \includegraphics[width=\linewidth]{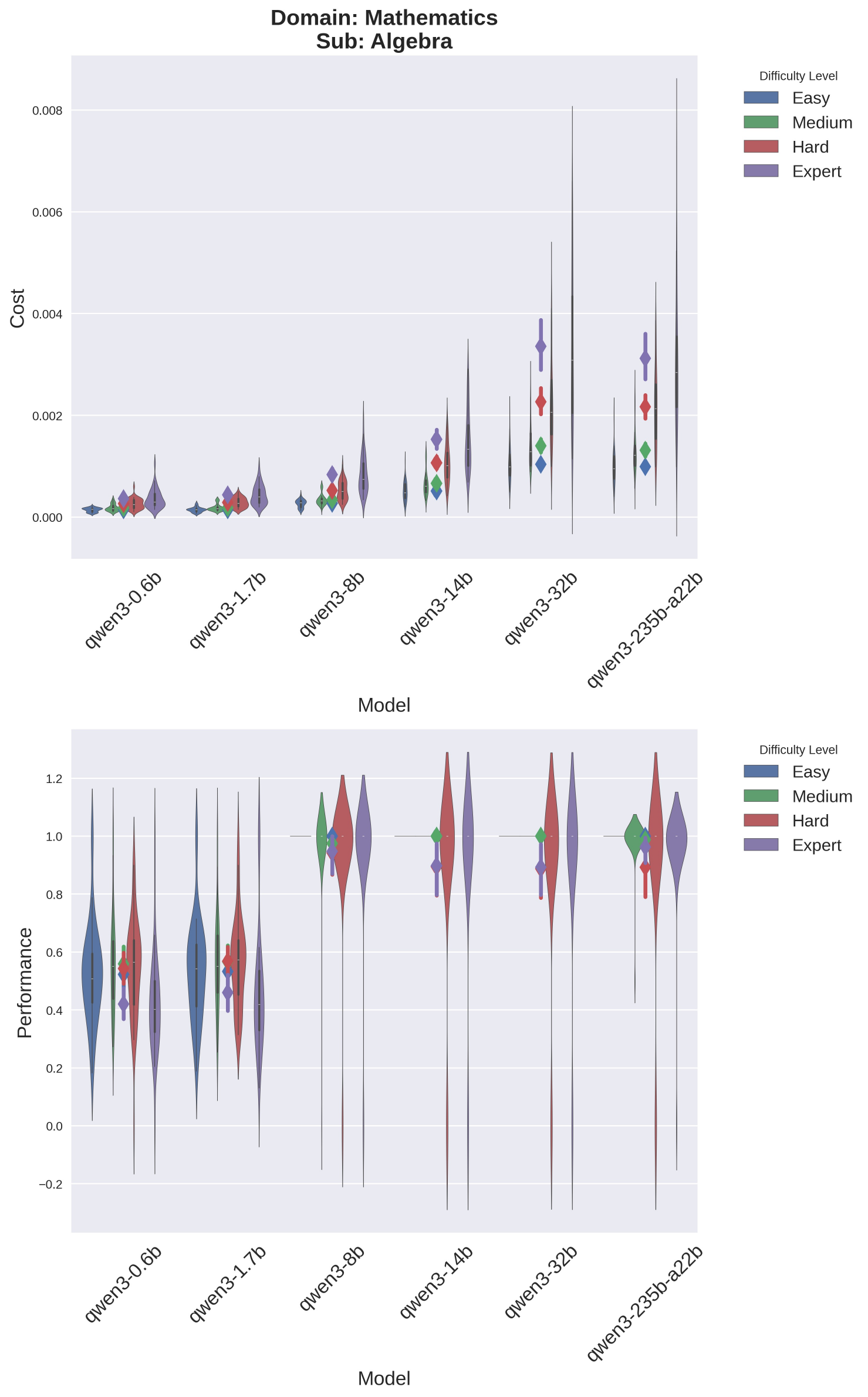}
  \end{subfigure}\hfill
  \begin{subfigure}[b]{0.33\linewidth}
    \centering
    \includegraphics[width=\linewidth]{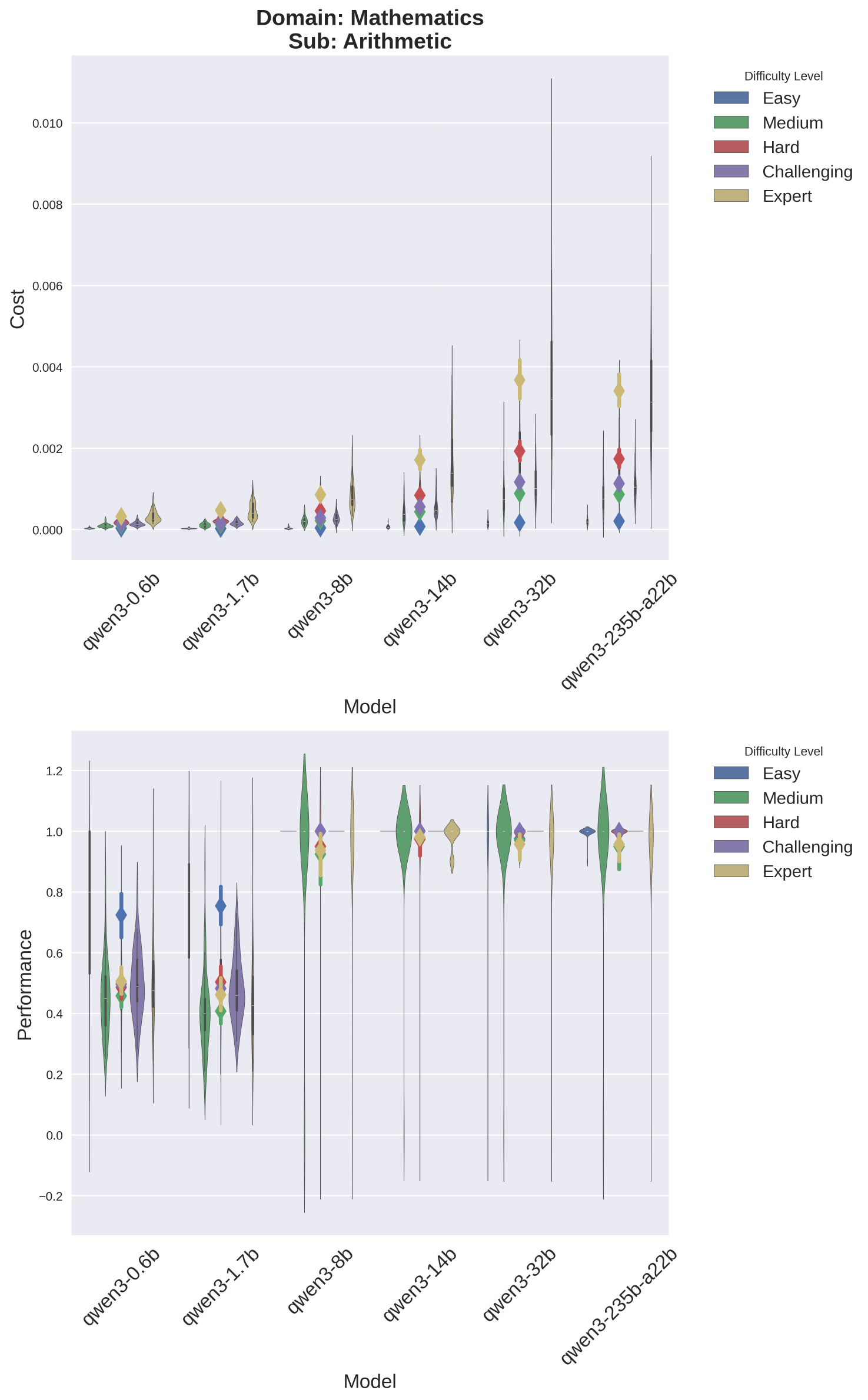}
  \end{subfigure}\hfill
  \begin{subfigure}[b]{0.33\linewidth}
    \centering
    \includegraphics[width=\linewidth]{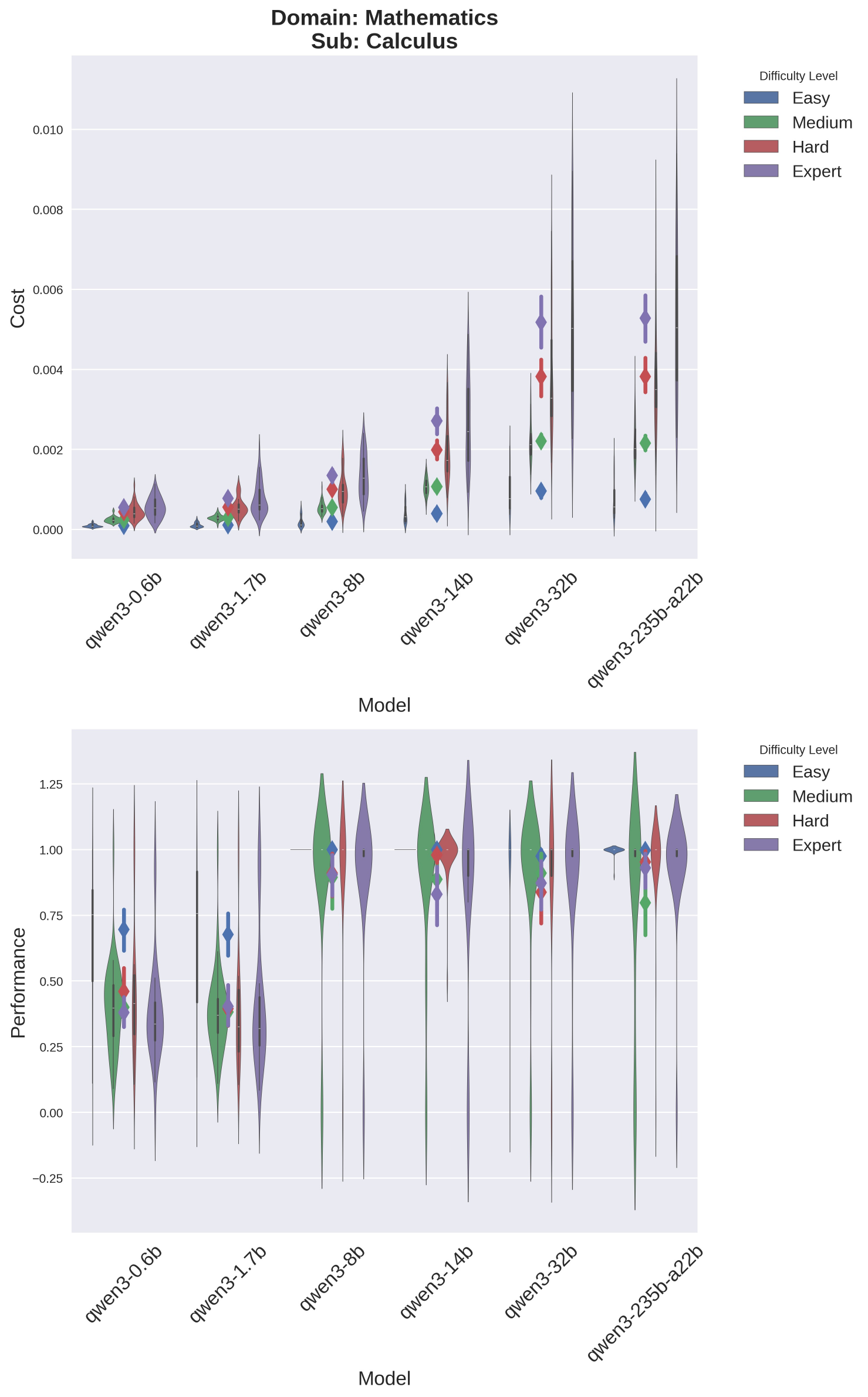}
  \end{subfigure}

  \caption{Visualization of cost and performance distributions for six Qwen-series models across representative subcategories of varying difficulty in the Mathematics domain, derived from the synthesized dataset $\mathcal{D}_{\text{syn}}$.}
  \label{fig:grid9math}
\end{figure*}
\subsection{More Ablation Studies}
\paragraph{Experiments on closed-source candidate LLMs.} Beyond the main experiment where we use the open-source Qwen 3 series as candidate LLMs, we also include supplementary results on closed-source models in  Table~\ref{tab:geminigroup}. The results show that TRouter achieves state-of-the-art performance in both cold-start and in-domain settings on this pool. Furthermore, when trained on the synthesized dataset $\mathcal{D}_{\text{syn}}$, TRouter substantially outperforms Adaptive LLM, providing additional evidence that our task-profile–guided data synthesis is effective for cold-start routing.
\paragraph{Ablation on LLMs for data synthesis.} As shown in Table~\ref{tab:maintraditionmetric}, instantiating our task-type synthesis framework with different LLMs (e.g., GPT-4.1, Gemini-2.5-Flash) yields consistent gains and both the task-profile–guided data synthesis and the task-type–aware routing (TRouter) framework outperform baselines across settings. These results indicate these frameworks' versatility with respect to the underlying synthesizer model.
\paragraph{Effect of using Manual label in TRouter.} TRouter infers task types via a recognition module. In the cold-start setting, we replace these task type predictions with ground-truth task labels at test time, denoted AgenticRouter in Table~\ref{tab:suppablation_component}. AgenticRouter yields further gains over TRouter, but requires expensive LLM-based task classification at inference. In contrast, TRouter dose not require this cost, offering a more efficient solution. 
\paragraph{Ablation study of  Assumptions (1).} We relax Assumption (1) in Sec. \ref{sec.3.3.1}, which posits performance is independent of the query given task type and model. Without this assumption. Hence, $p(r \mid q, m)$ can be re-formulated as follows:
\begin{equation}
\small
p(r \mid q, m) = \sum_{t \in \mathcal{T}} p(r \mid t, m,q)p(t \mid q).
\end{equation}
Results in Table~\ref{tab:suppablation_component} show that the original TRouter (with Assumption 1) consistently outperforms the variant without this assumption and is more robust across different user-preference settings, especially under the cold-start setting.  

Additionally, when releasing both assumption in Eq. \eqref{eq:3}, the resulting model is MetricRouter. However, as shown in Table \ref{tab:geminigroup}, Table \ref{tab:mainllmasjudge} and Table \ref{tab:maintraditionmetric}, MetricRouter performs worse than TRouter. Therefore, we suggest that accurately modeling the relationship between each individual query and its associated cost and performance is non-trivial, given the diversity of user queries and the dynamic nature of LLM outputs, especially for a lightweight router model with only a few hundred million parameters. Overfitting to query-specific information (as in MetricRouter) can lead to spurious correlations. While fine-tuning a large language model to capture such query-specific nuances may be feasible, it would drastically increase both latency and computational cost. As such, we conclude that task-type-guided prediction offers a more robust and efficient solution for routing in both cold-start and in-domain settings.
\paragraph{Effect of learning rate.} We vary the learning rate of TRouter from 1e-3 to 5e-5 in the cold-start setting and report the average results over five runs in Fig.~\ref{fig:lrrate}.  The results indicate that the routing utility peaks around a learning rate of  1e-4 and remains consistent as the learning rate is further decreased. Finally, we choose 1e-4 as the default learning rate in our main experiments.
\paragraph{Effect of Multi-task-type-label}
 We conduct an additional experiment using \texttt{GPT-4.1-2025-04-14} to annotate multi-label task type distributions for each question, in contrast to the single-task-type supervision used in the main setup. The task recognition module is then trained by minimizing the KL divergence between the predicted distribution and the ground-truth multi-label distribution. Specifically, binary label vectors (e.g., [0, 1, 1, 0, ...]) were converted into valid probability distributions using the SoftMax function. Results in Table \ref{tab:multitask_vs_single} show no significant performance difference between multi- and single-task-type supervision, likely because our original approach already incorporates probabilistic task assignment through marginalization over all task types via a learned distribution (Eq (4) and Eq (7) in the main paper), effectively capturing task ambiguity and uncertainty. 
 \begin{table*}[ht]
\centering
\caption{Results across three user preference settings in both cold-start and in-domain scenarios on an additional closed-source candidate LLM pool. For in-domain training, candidate LLM performance is obtained using \textbf{traditional metrics}. Bold indicates the best result for each user preference. The candidate pool consists of \texttt{gpt-5}, \texttt{gpt-5-mini}, \texttt{gpt-5-nano}, and \texttt{Doubao-Seed-1.6-Flash}.}
\label{tab:gptgroup}
\begin{adjustbox}{max width=\textwidth}
\begin{tabular}{ll
                *{3}{S[table-format=1.4]}
                *{3}{S[table-format=1.4]}
                *{3}{S[table-format=1.4]}
                S[table-format=1.4]}
\toprule
& & \multicolumn{3}{c}{Cost Preference} & \multicolumn{3}{c}{Balance Preference} & \multicolumn{3}{c}{Performance Preference} & {Utility Sum} \\
\cmidrule(lr){3-5} \cmidrule(lr){6-8} \cmidrule(lr){9-11}
Scenario & Method &
{Cost} & {Performance} & {Utility} &
{Cost} & {Performance} & {Utility} &
{Cost} & {Performance} & {Utility} &
{} \\
\midrule
Cold-start & Smallest LLM   & 0.0052 & 0.3987 & \textbf{0.0756} & 0.0052 & 0.3987 & 0.1967 & 0.0052 & 0.3987 & 0.3179 & 0.5902 \\
           & Largest LLM    & 0.4354 & 0.4573 & -0.2689 & 0.4354 & 0.4573 & 0.0109 & 0.4354 & 0.4573 & 0.2788 & 0.0328 \\
           & Adaptive LLM   & 0.0052 & 0.3987 & \textbf{0.0756} & 0.0223 & 0.3992 & 0.1883 & 0.4354 & 0.4573 & 0.2788 & 0.5426 \\
           & Ours           & 0.0116 & 0.4241 & 0.0755 & 0.0134 & 0.4280 & \textbf{0.2073} & 0.0577 & 0.4350 & \textbf{0.3365} & \textbf{0.6193} \\
\midrule
In-domain  & RouterDC       & 0.0099 & 0.4239 & \textbf{0.0787} & 0.0374 & 0.4448 & 0.2030 & 0.0374 & 0.4459 & 0.3492 & 0.6309 \\
           & MetricRoute    & 0.0100 & 0.4282 & 0.0776 & 0.0135 & 0.4307 & 0.2086 & 0.0586 & 0.4498 & 0.3482 & 0.6344 \\
           & Ours           & 0.0182 & 0.4451 & 0.0745 & 0.0080 & 0.4328 & \textbf{0.2124} & 0.0280 & 0.4577 & \textbf{0.3606} & \textbf{0.6474} \\
\midrule
Oracle     &                & 0.0085 & 0.4780 & 0.0888 & 0.0199 & 0.4996 & 0.2399 & 0.0323 & 0.5064 & 0.3987 & 0.7274 \\
\bottomrule
\end{tabular}
\end{adjustbox}
\end{table*}
\paragraph{Experiments on an additional closed-source candidate LLM pool.} 
Beyond the main experiments, we further evaluate TRouter on another closed-source candidate LLM pool to address the concern about model diversity. Specifically, in addition to the model groups used in the main paper, we construct a new pool comprising \texttt{gpt-5}, \texttt{gpt-5-mini}, \texttt{gpt-5-nano}, and \texttt{Doubao-Seed-1.6-Flash}, which span diverse capabilities and cost profiles. We conduct experiments on four datasets from Table~\ref{tab:modelstat}. The results are reported in Table~\ref{tab:gptgroup}. Similar to the findings in the main paper, TRouter achieves the best overall performance in both cold-start and in-domain settings on this candidate pool. In particular, under cold-start routing, TRouter consistently improves the overall utility compared with Adaptive LLM across different user preferences. In the in-domain setting, TRouter also achieves the highest utility sum, further demonstrating that our framework generalizes well across heterogeneous closed-source model families rather than relying on a specific model series.
\paragraph{Analysis of potential data leakage.} 
To assess the possibility of data leakage, we conduct two analyses on the test splits of GSM8K and SQuAD, using the training split of our synthesized data (30 examples per difficulty level) for comparison. First, we measure literal similarity by computing the cosine similarity between each test example and each synthetic example using embeddings from \texttt{all-MiniLM-L6-v2}. For every test sample, we retrieve the top-20 most similar synthetic examples and count cases with similarity equal to 1.0 (exact match) or at least 0.95 (high similarity). We find no exact or high-similarity matches on GSM8K, and only one such case on SQuAD. Second, under a stricter criterion, we further examine semantic equivalence, where two questions differ in wording but express the same intent. Specifically, we ask GPT-4.1 to judge whether each benchmark query is semantically equivalent to any of its top-20 retrieved synthetic examples from the first analysis. Under this criterion, no semantically equivalent case is found on GSM8K and only two cases are identified on SQuAD. These results suggest that direct overlap between the synthesized training data and benchmark test sets is extremely limited. Moreover, TRouter already achieves near-optimal routing performance with only 5--10 training examples per difficulty level, substantially below the full 30-shot setting used in Table~4, and it consistently outperforms the baselines in the cold-start setting shown in Tables~3 and~4. Both observations further mitigate the possibility that our results are driven by data leakage. We also note that the similarity-based filtering in our pipeline is only used to remove redundancy within the synthesized data itself, rather than to filter against the test splits.
\subsection{Supplementary Dateset}
\label{app:suppexe}
Beyond the four benchmark datasets in the main body, we further evaluate TRouter on MBPP~\citep{mbpp}, MWT’23, MedMCQA~\citep{medmcqa}, and LegalBench~\citep{guha2023legalbench}, covering four additional domains. We apply the same task‑profile–guided data synthesis pipeline and use pass@1 for MBPP, BLEU for MWT, and F1 for MedMCQA and LegalBench. All datasets are split into train/validation/test sets using a 5:1:4 ratio. As shown in Table~\ref{tab:tablesupp}, TRouter again achieves consistently superior performance in both cold-start and in-domain settings, validating the effectiveness of our method.
\subsection{Analysis of metric distribution on $\mathcal{D}_\text{syn}$.}
\label{anaonDsyn}
To examine the cost and performance distribution of models of varying scales on our generated dataset $\mathcal{D}_\text{syn}$, we visualize these metrics for six Qwen-series models across representative subcategories of varying difficulty within the Commonsense Knowledge, Creative Writing, and Mathematics domains, derived from the synthesized dataset $\mathcal{D}_\text{syn}$ in Figures \ref{fig:grid3_commonsese}, \ref{fig:grid9creative}, and \ref{fig:grid9math}. The results demonstrate that cost increases monotonically with both model scale and task difficulty, while performance improves with scale but exhibits diminishing returns and declines as difficulty increases, most notably in Mathematics. The 7B and 14B models constitute the price-performance frontier for most subcategories, whereas 32B and larger models provide distinctive value primarily on high-difficulty mathematics and advanced reasoning tasks. These observations support the motivation for our proposed task-type-aware router (TRouter) and task-profile-guided data synthesis, which faithfully reproduces complex, real-world cost trends with minimal expense compared to training routers on manually collected in-domain data. Additional analyses are presented below:
\paragraph{Cross-domain patterns}
Across all domains, cost increases steadily with scale while variance expands for larger models. Mathematics exhibits the steepest high-end cost growth, consistent with longer reasoning chains, while Creative Writing demonstrates the lowest costs at equivalent scales, with Commonsense falling between these extremes. Performance improves with scale across all domains, yet gains diminish beyond 32B parameters. Mathematics shows the highest scale sensitivity, with small models performing markedly worse while 14B and 32B models close the performance gap and can exceed larger models on easier subsets. Creative Writing reaches an early performance ceiling at higher difficulty levels, while Commonsense demonstrates steady, moderate improvements.
\paragraph{Same domain, different subcategories}
In the Commonsense domain, Category Membership \& Classification reaches saturation early with modest cost growth and stable improvements. Causal Reasoning \& Consequences requires greater computational resources, with clear advantages for larger models on Hard and Expert difficulty levels and extended cost distributions. Comparative \& Quantitative Reasoning falls between these patterns, exhibiting difficulty-driven performance drops more pronounced than those observed in classification tasks.

In Creative Writing, Character Creation and Genre or Style Evaluation show minimal performance gaps at Easy and Medium difficulty levels, diverging only at Hard and Expert levels while maintaining relatively low costs. Dialogue Generation demonstrates greater scale sensitivity, benefiting from larger models at higher difficulty levels while incurring moderately higher costs due to extended output lengths.

In Mathematics, Arithmetic benefits most directly from scaling, with 14B+ models maintaining strong median performance on Hard difficulty and exhibiting pronounced long-tailed cost distributions. Algebra displays greater variance and requires medium-to-large models for Hard and Expert difficulties. Calculus presents the steepest difficulty gradient, where even large models experience performance degradation and increased costs at Expert level.

\paragraph{Fixed domain and subcategory analysis}
When holding both domain and subcategory fixed, cost increases monotonically with model scale, with larger models displaying heavier-tailed distributions reflecting rare but computationally expensive samples. Performance improves with model scale, but the cost-performance ratio exhibits diminishing returns. The 1.7B models provide limited gains and demonstrate high sensitivity to difficulty variations. The 7B to 14B range represents the price-performance frontier with the largest marginal benefits. Models at 32B parameters and above continue to improve but with smaller gains, except for notable advantages on high-difficulty mathematics and complex reasoning tasks

\paragraph{Same subcategory, varying difficulty}
Within a given subcategory, cost increases monotonically from Easy to Expert difficulty levels, with the most pronounced increases observed in Mathematics and the most modest in Creative Writing. Performance exhibits a monotonic decline as difficulty increases, with the steepest degradation occurring in Mathematics, where only larger models maintain positive performance margins on Expert-level tasks. Commonsense tasks show moderate performance decline while continuing to benefit from scaling at higher difficulty levels. Creative Writing demonstrates minimal performance gaps through Hard difficulty, with a more pronounced decline appearing only at Expert level.

\paragraph{Cost–performance implications and methodological relevance}
The analysis reveals that 7B and 14B models deliver optimal unit-cost performance across most subcategories and difficulty levels, while 32B and larger models should be reserved for applications emphasizing high-difficulty Mathematics or advanced causal and quantitative reasoning tasks. These findings provide direct empirical justification for a task-type-aware routing strategy. Such a router would optimally allocate Easy or Medium-difficulty stylistic writing and classification tasks to 1.7B, 7B, or 14B models, escalating to 32B or larger models only when the task profile indicates reasoning-intensive workloads such as Mathematics-Calculus, Commonsense-Causal reasoning, or Dialogue Generation at Hard or Expert difficulty levels. This adaptive routing strategy reduces heavy-tail computational costs while preserving performance in domains where scaling demonstrates clear benefits (e.g., sometime due to the randomness, we need run LLM for serval times for a given query to obtain satisfactory results). 

Furthermore, the observed consistency between cost distributions, difficulty levels, and reasoning complexity substantiates the effectiveness of our task-profile-guided data synthesis approach. This methodology reproduces realistic cost escalation patterns and performance frontiers, enabling the development of benchmarking frameworks and routing policies that transfer effectively to complex real-world deployments while maintaining strong interpretability.

In summary, these empirical observations support the theoretical motivation underlying our proposed task-type-aware router (TRouter) and task-profile-guided data synthesis methodology, which faithfully reproduces complex real-world cost trends with minimal computational overhead compared to training routers on manually collected in-domain data.
\subsection{Interpretability Analysis}
\label{app:interpretableana}
After being trained on $D_{\text{train}}$, TRoute achieved a top-1 task type prediction accuracy of around 30\% on the test set, with a task space size of $|\mathcal{T}| = 447$ while the top-3 accuracy reaches 50\% and  top-5 accuracy reaches 70\%.   These results indicate that the latent task-type modeling is able to capture meaningful and discriminative task semantics.  Moreover, experimental analysis reveals that most prediction errors originated from the Alpaca dataset, where the top-1 accuracy dropped to just 0.03.

Owing to the interpretability of TRouter, we further investigated the failure cases on the Alpaca dataset by visualizing the top-5 predictions alongside the ground truth in Table ~\ref{tab:failedcaseonapa}. The results show that a single test query may correspond to multiple task types. For example, the first case could belong to several types within Information Retrieval. However, since our model uses a weighted combination of task types instead of an argmax selection as defined in Eq. \eqref{eq:4}, this ambiguity could be mitigated, still leading effective routing decision.

A key advantage of TRouter over black-box embedding-based routers is its interpretable taxonomy structure, which directly facilitates scalability and maintenance through 1) Node Offloading. Users/more advanced LLMs can easily inspect the generated taxonomy. If certain sub-categories are irrelevant to their specific deployment environment, they can simply remove those nodes. 2) Bias Correction. For example, users/more advanced LLMs can observe the routing distribution across the taxonomy. If they detect distributional shift (e.g., a specific node consistently routing to an overly expensive model due to wrong performance prediction), they can manually refine this specific node's cost/performance parameters. 3) Incremental Updates: When new tasks emerge, the hierarchical nature of our method allows for incremental expansion. Users can generate a new branch for the new task and merge it into the existing tree, rather than re-synthesizing the entire taxonomy from scratch. Overall, TRouter has the potential to evolve further in future applications. However, such discussions are beyond the scope of this work, which focuses specifically on the cold-start scenario. We leave these directions for future exploration.
\begin{table*}[ht]
\centering
\caption{Comparison between single-task-type and multi-task-type training for task recognition.}
\label{tab:multitask_vs_single}
\begin{adjustbox}{max width=\textwidth}
\begin{tabular}{l
                *{3}{S[table-format=1.4]}
                *{3}{S[table-format=1.4]}
                *{3}{S[table-format=1.4]}
                S[table-format=1.4]}
\toprule
& \multicolumn{3}{c}{Cost First} & \multicolumn{3}{c}{Balance} & \multicolumn{3}{c}{Performance} & {Utility Sum} \\
\cmidrule(lr){2-4} \cmidrule(lr){5-7} \cmidrule(lr){8-10}
\textbf{Task-type for training} &
{Cost} & {Perf} & {Utility} &
{Cost} & {Perf} & {Utility} &
{Cost} & {Perf} & {Utility} &
{} \\
\midrule
Multi-task-type       & 0.0414 & 0.4249 & 0.0519 & 0.0409 & 0.4308 & 0.1950 & 0.0678 & 0.4401 & 0.3385 & 0.5853 \\
Single-task-type (Ours) & 0.0393 & 0.4165 & 0.0518 & 0.0477 & 0.4424 & 0.1974 & 0.0751 & 0.4479 & 0.3433 & 0.5925 \\
\bottomrule
\end{tabular}
\end{adjustbox}
\end{table*}

\begin{table*}[ht]
\centering
\caption{Comparison between TRouter trained on LLM-generated data and on manually labeled data.}
\label{tab:comllmmann}
\begin{adjustbox}{max width=\textwidth}
\begin{tabular}{l
                *{3}{S[table-format=1.4]}
                *{3}{S[table-format=1.4]}
                *{3}{S[table-format=1.4]}
                S[table-format=1.4]}
\toprule
& \multicolumn{3}{c}{Cost First} & \multicolumn{3}{c}{Balance} & \multicolumn{3}{c}{Performance} & {Utility Sum} \\
\cmidrule(lr){2-4} \cmidrule(lr){5-7} \cmidrule(lr){8-10}
Method &
{Cost} & {Perf} & {Utility} &
{Cost} & {Perf} & {Utility} &
{Cost} & {Perf} & {Utility} &
{} \\
\midrule
Ours            & 0.0345 & 0.3098 & 0.0343 & 0.0640 & 0.4261 & 0.1810 & 0.0684 & 0.4224 & 0.3243 & 0.5397 \\
Manual Labeled  & 0.0332 & 0.3115 & 0.0357 & 0.0601 & 0.4298 & 0.1849 & 0.0710 & 0.4201 & 0.3219 & 0.5425 \\
\bottomrule
\end{tabular}
\end{adjustbox}
\end{table*}
\section{Quantitative Analysis of the Quality of Generated QA Pairs}
\label{appendix:quaqualityofqapairs}
In the cold-start setting, we propose a multi-level task-profile-guided data synthesis framework that requires only domain descriptions to guide LLMs in generating synthetic QA datasets. This approach eliminates the need for manual data collection and annotation. Although prior work shows that low-diversity data offers limited benefit~\citep{yang2025measuring}, and poor-quality synthetic data can harm models or lead to collapse~\citep{shumailov2024ai}, LLM routing is more robust against these limitations of data synthesis because its objective is to capture cost–performance tradeoffs across models and heterogeneous queries, rather than to execute tasks precisely, the strict quality constraints on training data can be moderately relaxed.  Given the large volume of generated data (e.g., 17,880 QA pairs from GPT-4.1-2025-04-14) and the difficulty of some questions, especially in domains like mathematics and commonsense reasoning, which are challenging even for annotators with STEM master's or PhD degrees, it is infeasible to conduct a fully manual quantitative analysis of data quality and diversity. Therefore, to assess the quality of the generated QA pairs, we evaluate the answer accuracy and QA diversity through the following experiments.

\paragraph{Accuracy} We focus on three domains, involving Mathematics, Creative Writing, and Commonsense Knowledge. Since answer accuracy is typically correlated with task difficulty, we categorize each subcategory node into three difficulty levels, including easy, medium, and hard. If a node contains more than three difficulty levels, we map all non-extreme levels to medium. For each domain, we sample 160 easy, 160$\times$N medium, and 160 hard questions (where N is the number of difficulty node excepting easy and hard nodes). For commonsense knowledge, due to the breadth of topics and inefficiency of manual retrieval, we use Perplexity for retrieval, followed by human verification. The results show answer accuracy is consistently high across difficulty levels, nearly 100\% for easy questions, 97.16\% for medium, and 94.5\% for hard. Additionally, the accuracy at hard level varies by domain, with hard questions in Mathematics achieving 92.5\% and those in Commonsense Knowledge reaching 97.5\%. These findings suggest that many of the QA pairs may source from the training data of large models and can be effectively retrieved via generation. This may explain the high accuracy across diverse domains and even at difficulty levels.

\paragraph{Diversity} We evaluate diversity at both the subcategory (103 nodes) and difficulty (447 nodes) levels, finding no semantically redundant nodes. While some conceptual similarities exist (e.g.,Long-Form Summarization from Long-context Understanding and Main Idea \& Summarization from Reading Comprehension domain), they arise from different emphases across domains or contexts. For instance, the former focuses on long-context compression, while the latter targets reading comprehension.We also examine intra-node diversity by analyzing 30 difficulty nodes from the Progr amming and Commonsense Knowledge domains. Since we remove semantically similar QA pairs during generation, no near-duplicate questions remain . Moreover, each difficulty node covers a diverse range of topics. For example, one programming node covers 12 core topics, including Python multiple inheritance, asynchronous testing, and documentation for complex/generic classes. However, if we consider semantically equivalent questions with different phrasing as redundant, the redundancy rate varies significantly by domain. In the programming domain, semantic redundancy is minimal, whereas in commonsense domains it is approximately 10\%.

\section{Comparison between Synthesized and Manually Labeled Dataset} 
As outlined in Sec. \ref{sec.3.2}, our data synthesis pipeline comprises three components, including the Task Type Generator, Task Type Quality Evaluator, and Question-Answer Pair Generator. Despite multiple attempts, three PhD-level researchers could not produce a diverse and high-quality set of task types, even within their areas of expertise (e.g., programming, creative writing). In contrast, LLMs generate a broad and diverse range of task types efficiently. This is likely because annotating task types requires comprehensive knowledge of domain-specific concepts and the ability to distinguish between varying difficulty levels. Accordingly, we adopt task types generated by LLMs.Moreover, manual construction of a large volume of QA pairs is also impractical. Annotators often lack the expertise to cover specialized topics such as external APIs, mocking, multiple inheritance, or mixed async/sync programming in programming domain. Therefore, we use LLMs to generate candidate QA pairs and then manually verify these data. From these, we select five QA pairs per difficulty node, resulting in a curated dataset of 2,235 QA pairs. For cold-start evaluation, we split the data into 6:4 for training and validation.

As shown in Table~\ref{tab:comllmmann}, performance differences between synthetic and manually labeled datasets are minimal. We argue that is this outcome is expected, as training on data drawn from the same distribution and evaluated under consistent protocols (i.e., using LLM-as-a-judge for both training and test sets) yields more reliable results, as evidenced in Table~\ref{tab:mainllmasjudge}.

\section{Cost Breakdown Analysis}
\label{app:costbreakdownanalysis}
For synthetic data generation, we employ our multi-level task-profile-guided synthesis framework powered by \texttt{GPT-4.1-2025-04-14}. As detailed in Appendix \ref{app:imple}, generating 17,880 QA pairs incurs a total cost of approximately \$14.34. In contrast, manual annotation is both economically and practically infeasible. Even PhD-level researchers encountered difficulties in producing diverse task taxonomies and lacked expertise in specialized domains (e.g., complex coding scenarios involving external API mocking).

To estimate manual annotation costs, we assume: 15 minutes per hard QA pair, 3 minutes per medium, and 1 minute per easy question. Unlabeled items are classified as medium difficulty. The dataset includes 103$\times$40 easy, 241$\times$40 medium, and 103$\times$40 hard questions. At a labor rate of \$30 per 8-hour day, the total estimated cost exceeds \$6,000.

In-domain data collection is also impractical in a cold-start setting, as real user queries are unavailable and recruiting test users would incur costs potentially in the tens of thousands of dollars. Consequently, LLM-based automatic data synthesis is a more viable approach for training task routers under cold-start conditions.

\clearpage
\begin{table*}[t]
\centering
\captionsetup{width=\linewidth}
\small
\setlength{\tabcolsep}{6pt}
\renewcommand{\arraystretch}{1.15}
\begin{tabular}{p{14.5cm}}
\toprule
\begin{minipage}[t]{\linewidth}
\small
\begingroup
\ttfamily
\obeylines
\textbf{Example 1}:
query: Create a list of 4 tips to become a better public speaker. 
True Task Type: {'domain': 'Dialogue \& Communication', 'subcategory': 'Interactive Tutoring \& Socratic Dialogue', 'difficulty': 'Medium'}    
Definition: Providing brief, step-based guidance or explanation for standard single-step problems with clear solutions.     
Top-5 Predictions:
1. Easy (Scientific Reasoning - Scientific Explanation) Possibility: 36.72\%                                             Definition: Explaining well-known, fundamental scientific concepts or phenomena using basic language and without requiring prior specialized knowledge.               
2. Easy (Reading Comprehension - Fact \& Detail Retrieval) Possibility: 19.85\%                                          Definition:  Identify and extract a single, explicitly stated fact or detail from a sentence, with a direct, unambiguous answer (typically a word or short phrase).                
3. Easy (Information Retrieval - Fact Lookup) Possibility: 10.54\%                                                       
Definition: Retrieval of a single, widely-known fact stated explicitly in authoritative sources. 
4. Easy (Information Retrieval - Definition/Explanation Retrieval) Possibility: 3.71\%                                   
Definition: Retrieving straightforward definitions or explanations of widely recognized terms or concepts with minimal ambiguity.                        
5. Moderate (Information Retrieval - Fact Lookup) Possibility: 3.58\%      
Definition: Retrieval of a less common fact or fact requiring disambiguation (e.g., multiple entities, timeframes, or similar-sounding terms). 
\vspace{15pt}
\textbf{Example 2}:
query: Choose the correct pairing for the given words.Drawing, music. 
True Task Type:  {'domain': 'Scientific Reasoning', 'subcategory': 'Concept Comparison', 'difficulty': 'Moderate'}  
Definition: Comparison of related or similar scientific concepts that may have subtle distinctions or overlapping features, requiring more detailed analysis.

Top-5 Predictions: 
1. Easy (Scientific Reasoning - Scientific Explanation) Possibility: 44.46\%                                             
Definition:Retrieval of a single, widely-known fact stated explicitly in authoritative sources.                          
2. Easy (Reading Comprehension - Fact \& Detail Retrieval) Possibility: 26.48\%                                          Definition: Listing well-known, static items from a single, unambiguous category with no need for filtering or reasoning.                          
3. Easy (Scientific Reasoning - Scientific Explanation) Possibility: 14.69\%                                             Definition: Explaining well-known, fundamental scientific concepts or phenomena using basic language and without requiring prior specialized knowledge.                             
4. Medium (Information Retrieval - List Generation)   Possibility: 4.08\%                                   
Definition: Listing items from a category with simple, explicit criteria or filters, requiring only basic fact retrieval and minimal reasoning.                          
5. Moderate (Information Retrieval - Fact Lookup) Possibility: 2.75\%   
Definition: Assessing the physical possibility of straightforward, everyday actions or events that rely on well-known physical laws and typical human abilities.    
\endgroup
\end{minipage}
\\
\bottomrule
\end{tabular} 
\caption{Failure Cases of the task recognition module of TRouter.} 
\label{tab:failedcaseonapa} 
\end{table*}

\clearpage
\begin{table*}[t]
\centering
\captionsetup{width=\linewidth} 
\small
\setlength{\tabcolsep}{6pt}
\renewcommand{\arraystretch}{1.15}
\begin{tabular}{p{14.5cm}}
\toprule
\begin{minipage}[t]{\linewidth}
\small
\begingroup
\ttfamily
\obeylines
Prompt of LLM-as-Judge
You are an expert evaluator. Your task is to score the quality and correctness of a model-generated answer to a given question, using a reference answer as the gold standard.

You will be given:
- QUESTION
- REFERENCE ANSWER (correct)
- MODEL ANSWER (to evaluate)

Your goal is to assign a score between **0.0 and 1.0**, where:
- 1.0 = Fully correct and semantically equivalent to the reference.
- 0.5 = Partially correct or incomplete
- 0.0 = Completely incorrect, irrelevant, or nonsensical

Respond with **only the numeric score**, nothing else.

---

QUESTION:
\{question\}

REFERENCE ANSWER:
\{reference\_answer\}

MODEL ANSWER:
\{response\}.
Your score:
\endgroup
\end{minipage}
\\
\bottomrule
\end{tabular}
\caption{Prompt of LLM-as-Judge. The \{question\}, \{reference\_answer\}, and \{response\} will be instantiated by the question, answer, and response from candidate LLMs.}
\label{tab:evaluator-prompt}
\end{table*}

\begin{table*}[t]
\centering
\captionsetup{width=\linewidth}
\small
\setlength{\tabcolsep}{6pt}
\renewcommand{\arraystretch}{1.15}
\begin{tabular}{p{14.5cm}}
\toprule
\begin{minipage}[t]{\linewidth}
\small
\begingroup
\ttfamily
\obeylines
TaskTypeGenSystemPrompt = """
You are a knowledge graph construction expert. Your task is to help build a hierarchical classification system for an intelligent model router called AgenticiRouter.

AgenticiRouter selects the most appropriate Large Language Model (LLM) based on a user's query and preferences (e.g., performance, cost-efficiency, latency). To support this, we are building a multi-level taxonomy of query types.

Each query is classified into a three-level node path:

Domain — broad task area
Subcategory — specific task type
Difficulty Level — complexity level within that task type
"""
\endgroup
\end{minipage}
\\
\bottomrule
\end{tabular} 
\caption{System Prompt of Task type Generation.} 
\label{tab:system-prompt} 
\end{table*} 

\begin{table*}[t]
\centering
\captionsetup{width=\linewidth}
\small
\setlength{\tabcolsep}{6pt}
\renewcommand{\arraystretch}{1.15}
\begin{tabular}{p{14.5cm}}
\toprule
\begin{minipage}[t]{\linewidth}
\small
\begingroup
\ttfamily
\obeylines
DomainNodeRule =   """
- Domains must be general yet semantically distinct.
- Avoid overlapping or ambiguous categories.
- Think of areas commonly found in LLM benchmarks (e.g., MMLU, AGIEval, CMMLU) and real-world applications.
- These will serve as the top-level nodes of the taxonomy.
- Only up to \{max\_new\_domain\_nodes\} new domains can be proposed.
"""
\vspace{6pt} 

SubcategoryNodeRule = """
- Avoid overlap between subcategories.
- Each subcategory should represent a common type of user query that can be grouped under this domain.
- Subcategories should be general enough to cover many user queries but specific enough to guide model selection.
- Only up to \{max\_subcategory\_nodes\} new Subcategory Nodes can be proposed for each domain node.
"""
\vspace{6pt} 

DifficultyLevelNodeRule = """
- Levels must be ordered from easiest to hardest.
- Levels should reflect increasing reasoning complexity, token usage, or LLM capability required.
- Levels should be mutually exclusive — no query should belong to more than one level.
- Levels should be collectively exhaustive — all queries in this subcategory must be covered.
- Avoid generic differences like “longer text” unless it reflects actual difficulty in reasoning or generation.
- Only up to \{max\_difficulty\_level\_nodes\} Difficulty Level Nodes can be proposed for each Subcategory Node.
"""
\endgroup
\end{minipage}
\\
\bottomrule
\end{tabular} 
\caption{Rules of nodes generation at three different levels, which will be instantiated to corresponding node generation prompts. \{max\_new\_domain\_nodes\}, \{max\_subcategory\_nodes\}, and \{max\_difficulty\_level\_nodes\} will be set as 4, 10, 5 in our work.} 
\label{tab:noderule} 
\end{table*} 

\begin{table*}[t]
\centering
\captionsetup{width=\linewidth}
\small
\setlength{\tabcolsep}{6pt}
\renewcommand{\arraystretch}{1.15}
\begin{tabular}{p{14.5cm}}
\toprule
\begin{minipage}[t]{\linewidth}
\small
\begingroup
\ttfamily
\obeylines
DomainNodePrompt = TaskTypeGenSystemPrompt + """
**Step 1: Generate Domain Nodes**

Generate a list of distinct, non-overlapping **Domain** categories that represent broad types of user tasks.
\vspace{6pt} 
We have already defined the following six initial Domains:

- Mathematics  
- Creative Writing  
- Commonsense Knowledge 
- Programming 
- Long-context Understanding 
- Reading Comprehension
\vspace{6pt} 
Please propose **\{max\_new\_domain\_nodes\}  new high-level Domains** that:

**Rules:**

\{DomainNodeRule\}
\vspace{6pt} 
**Output Format:**

For each proposed Domain:
- Name the Domain
- Provide a one-sentence definition
- Include a real-world example query
\vspace{6pt} 
**Example:**
<node begin>
Domain: Mathematics
Definition: Covers quantitative problem-solving tasks involving numbers, equations, logic, or formal systems, including arithmetic, algebra, calculus, and more.
Example: What is the derivative of sin(x)?

Domain: Creative Writing
Definition: Involves imaginative or artistic language generation tasks such as writing poems, stories, scripts, or creative descriptions.
Example: Write a short story about a robot who learns to paint.
</node end>
\vspace{6pt} 
**Output:**
"""
\endgroup
\end{minipage}
\\
\bottomrule
\end{tabular} 
\caption{Prompt of domain domain expansion. \{max\_new\_domain\_nodes\} will be set as 4 in our work. \{DomainNodeRule\} will be replaced using rules as shown in the Table. \ref{tab:noderule}.} 
\label{tab:domainnodegenprompt} 
\end{table*}

\begin{table*}[t]
\centering
\captionsetup{width=\linewidth}
\small
\setlength{\tabcolsep}{6pt}
\renewcommand{\arraystretch}{1.15}
\begin{tabular}{p{14.5cm}}
\toprule
\begin{minipage}[t]{\linewidth}
\small
\begingroup
\ttfamily
\obeylines
SubcategoryNodePrompt = TaskTypeGenSystemPrompt + """
**Step 2: Generate Subcategories for a Given Domain**

The current Domain is: \{domain\_name\}
Domain Definition: \{domain\_definition\}
Domain Example: \{domain\_example\}

Please generate a list of fine-grained Subcategories Nodes that represent specific types of tasks or problem types within this domain.

Please propose **up to \{max\_subcategory\_nodes\} new Subcategory Nodes** for each domain node that:

**Rules:**

\{SubcategoryNodeRule\}
\vspace{6pt} 
**Output Format:**
For each proposed Domain:
- Name the Subcategory
- Provide a one-sentence definition
- Include a real-world example query
\vspace{6pt} 

Example: 

Assume the Domain is:

Domain\_name: Mathematics
Domain Definition: Covers quantitative problem-solving tasks involving numbers, equations, logic, or formal systems, including arithmetic, algebra, calculus, and more.
Domain Example: What is the derivative of sin(x)?

Then the Subcategory Nodes are:

<node begin>
Subcategory: Arithmetic
Definition: Covers basic arithmetic operations and problem-solving involving numbers, including addition, subtraction, multiplication, and division.
Example: What is the sum of 123 and 456?

Subcategory: Algebra
Definition: Involves solving equations and expressions using variables and algebraic manipulation.
Example: Solve for x in the equation 2x + 3 = 11.

...
</node end>
\vspace{6pt} 
**Output:**:
"""

\endgroup
\end{minipage}
\\
\bottomrule
\end{tabular} 
\caption{Prompt of subcategory node generation. \{max\_subcategory\_nodes\} will be set as 10 in our work. \{SubcategoryNodeRule\} will be replaced using rules as shown in the Table. \ref{tab:noderule}.  \{domain\_name\}, \{domain\_definition\}, and  
\{domain\_example\} will be replaced by the corresponding domain information.} 
\label{tab:subcategorynodegenprompt} 
\end{table*}

\begin{table*}[t]
\centering
\captionsetup{width=\linewidth}
\small
\setlength{\tabcolsep}{6pt}
\renewcommand{\arraystretch}{1.15}
\begin{tabular}{p{14.5cm}}
\toprule
\begin{minipage}[t]{\linewidth}
\small
\begingroup
\ttfamily
\obeylines
DifficultyNodePrompt = TaskTypeGenSystemPrompt  + """
Step 3: Define Difficulty Levels for a Given Subcategory

You are now working on level 3: **Difficulty Level**. This level captures how task complexity varies **within a specific Subcategory**.
\vspace{6pt} 
Please utilize a fixed global scale (e.g., Easy / Medium / Hard for three-level difficulty) as the short name of each level.  
However, the Definition of each level should be customized based on the nature of the specific Subcategory.

**Input Subcategory Information:**
- Domain: \{domain\_name\}  
- Subcategory: \{subcategory\_name\}  
- Subcategory Definition: \{subcategory\_definition\}  
- Subcategory Example Query: \{subcategory\_example\}
\vspace{6pt} 
Please propose **up to \{max\_difficulty\_level\_nodes\} Difficulty Level Nodes** that:

**Rules:**

\{DifficultyLevelNodeRule\}
\vspace{6pt} 
**Output Format:**

For each proposed level:
- Name the Level
- Provide a one-sentence definition
- Include a real-world example query
\vspace{6pt} 
**Example**(Subcategory: Code Debugging): 
Assume the Subcategory is  Code Debugging.
Definition: Identifying and fixing errors in code snippets written in common programming languages.  

Then the Subcategory Nodes are:
<node begin>
Level: Basic
Definition: Single-line or syntax-only bugs in short, self-contained functions.
Example: Fix the indentation error in this 5-line Python function.

Level: Intermediate
Definition: Logic bugs in medium-length code that require control flow analysis.
Example: Fix the loop condition that causes an infinite loop in this JavaScript function.

Level: Advanced
Definition: Bugs across multiple functions, handling of edge cases, or requiring domain-specific knowledge.
Example: Fix the bug in this Flask app that breaks when uploading empty files.

Level: Expert
Definition: Deep reasoning over large codebases, concurrency, or memory management.
Example: Fix the deadlock issue in this multi-threaded C++ program handling file writes.
</node end>
\vspace{6pt} 
**Output:**
"""
\endgroup
\end{minipage}
\\
\bottomrule
\end{tabular} 
\caption{Prompt of difficulty node generation. \{max\_difficulty\_level\_nodes\} will be set as 5 in our work. \{DifficultyLevelNodeRule\} will be replaced using rules as shown in the Table. \ref{tab:noderule}.  \{domain\_name\}, \{subcategory\_name\},\{subcategory\_definition\}  and  
\{subcategory\_example\} will be replaced by the corresponding domain and subcategory information.} 
\label{tab:difficultydegenprompt} 
\end{table*}

\begin{table*}[t]
\centering
\captionsetup{width=\linewidth}
\small
\setlength{\tabcolsep}{6pt}
\renewcommand{\arraystretch}{1.15}
\begin{tabular}{p{14.5cm}}
\toprule
\begin{minipage}[t]{\linewidth}
\small
\begingroup
\ttfamily
\obeylines
NodeRevisePrompt =  TaskTypeGenSystemPrompt + """
You are given a **candidate \{node\_name\} set** for review. Your job is to:

1. Evaluate whether this candidate \{node\_name\} set needs improvement by checking how well it adheres to the provided generation rules..

2. If improvement is needed, generate a revised and higher-quality version of the \{node\_name\} set that better satisfies the rules and supports downstream LLM routing decisions.

Current Candidate \{node\_name\} Set:

\{candidate\_node\_set\}

**Node Generation Rules:**

\{node\_gen\_rules\}

**Output Format:**
<justification>
Explain whether the current \{node\_name\} set is flawed or could be improved. Mention overlap, vagueness, gaps, etc.
</justification>

<revision node begin>
\{node\_name\_short\}: node name
Definition: One-sentence definition  
Example: Example query or task

\{node\_name\_short\}: node name
Definition: One-sentence definition  
Example: Example query or task
</revision node end>

**Output:**"""
\vspace{12pt}
NodeSetChoicePrompt = TaskTypeGenSystemPrompt + """

Your Task: You are given **two candidate \{node\_name\} sets**, labeled **Set A** and **Set B**. Your job is to:

1. Compare both sets based on how well they follow the generation rules.  
2. Select the better set — the one that provides more clarity, distinctiveness, usefulness, and alignment with routing goals.
3. Justify your choice in detail.

**Node Generation Rules:**

\{node\_gen\_rules\}

**Candidate Sets:**

Set A:

\{candidate\_node\_set\_a\}

Set B:

\{candidate\_node\_set\_b\}

**Output Format:**
<justification>
Explain why one set is better than the other. Reference the rules. Mention clarity, distinctiveness, coverage, usefulness, etc.
</justification>
<preferred set>
Set A / Set B
</preferred set>

**Output:**
"""
\endgroup
\end{minipage}
\\
\bottomrule
\end{tabular} 
\caption{Prompt of difficulty task type quality evaluator.  \{node\_name\} represents the level of task type trees, including domain, subcategory, and difficulty. \{node\_gen\_rules\} will be replaced using rules corresponding to \{node\_name\} level, as shown in the Table. \ref{tab:noderule}.  \{candidate\_node\_set\_a\} and \{candidate\_node\_set\_b\} will be replaced by the original and revised node set.} 
\label{tab:tasktypequalityevaluator} 
\end{table*} 

\begin{table*}[t]
\centering
\captionsetup{width=\linewidth}
\small
\setlength{\tabcolsep}{6pt}
\renewcommand{\arraystretch}{1.15}
\begin{tabular}{p{14.5cm}}
\toprule
\begin{minipage}[t]{\linewidth}
\small
\begingroup
\ttfamily
\obeylines
QAGenPrompts = """
You are a helpful assistant that generates high-quality question-answer pairs. 
You must respond with a specific format using markers to structure your output.
\vspace{6pt}
IMPORTANT: Your response must follow this exact format:
\vspace{6pt}
<qa\_pairs begin>
Q: What is 2+2?
A: 2+2 equals 4

Q: What is the capital of France?
A: The capital of France is Paris

Q: How does photosynthesis work?
A: Photosynthesis is the process by which plants convert sunlight into energy.
</qa\_pairs end>
\vspace{6pt}
Rules:
- Start with <qa\_pairs begin> and end with </qa\_pairs end>
- Each QA pair should be separated by a blank line
- Use "Q:" for questions and "A:" for answers
- Ensure questions are clear, relevant, and the answers are accurate and comprehensive""" + 
\{TaskProfile\} +  """Please generate \{question\_num\} different question-answer pairs according to all the above specifications.
    The questions should be clear, relevant, and the answers should be comprehensive and accurate.
    Focus on creating diverse questions that cover different aspects of the topic."""

DomainNodeProfile = """
Task Domain: \{domain\_name\} 
Domain Definition: 
\{domain\_definition\}"""
\vspace{12pt}
SubcategoryNodeProfile =  """
Task Domain: \{domain\_name\}
Domain Definition: \{domain\_definition\} 
Task Subcategory: \{subcategory\_name\} 
Subcategory Definition: \{subcategory\_definition\}"""
\vspace{12pt}
DifficultyNodeProfile = """
Task Domain: \{domain\_name\}
Domain Definition: \{domain\_definition\}
Task Subcategory: \{subcategory\_name\}
Subcategory Definition: \{subcategory\_definition\}
Task Difficulty: \{difficulty\_name\}
Difficulty Definition: \{difficulty\_definition\}"""
\endgroup
\end{minipage}
\\
\bottomrule
\end{tabular} 
\caption{Prompt of question generator using task profile of different levels.  \{question\_num\}  will be set as 8 in our work. Other variable with \{ and \} will be replaced with the information of different task types and their parent nodes.} 
\label{tab:questiongeneratorr} 
\end{table*} 

\label{sec:appendix}

\end{document}